\documentclass[letterpaper]{article}
\usepackage{uai2018}
\usepackage[margin=1in]{geometry}

\usepackage{times}
\usepackage{graphicx}
\usepackage{subcaption}
\usepackage{amssymb}
\usepackage{amsmath}
\usepackage{amsthm}
\usepackage{booktabs} 
\usepackage{tikz}
\usepackage{bbm}
\usepackage{multirow}
\usepackage{comment}
\usepackage{natbib}
\usepackage{algorithmic}
\usepackage{algorithm}
\usepackage{hyperref}

\usepackage{xcolor}
\definecolor{dark-red}{rgb}{0.4,0.15,0.15}
\definecolor{dark-blue}{rgb}{0.15,0.15,0.4}
\definecolor{medium-blue}{rgb}{0,0,0.5}
\hypersetup{
   colorlinks, linkcolor={dark-blue},
   citecolor={dark-blue}, urlcolor={medium-blue}
}

\usepackage{comment}

\title{Averaging Weights Leads to Wider Optima and Better Generalization}

\author{
  Pavel Izmailov\thanks{\text{ }\text{ }Equal contribution.}\, $^1$ 
  \quad Dmitrii Podoprikhin$^*$$^{2,3}$ 
  \quad Timur Garipov$^*$$^{4,5}$ 
  \quad Dmitry Vetrov$^{2,3}$ 
  \quad Andrew Gordon Wilson$^1$ \\ 
  $^1$Cornell University, 
  $^2$Higher School of Economics, 
  $^3$Samsung-HSE Laboratory, \\
  $^4$Samsung AI Center in Moscow, 
  $^5$Lomonosov Moscow State University}

\begin{document}

\maketitle

\begin{abstract}

Deep neural networks are typically trained by optimizing a loss function with an
SGD variant, in conjunction with a decaying learning rate, until convergence.  We show that 
simple averaging of multiple points along the trajectory of SGD, with a cyclical
or constant learning rate, leads to better generalization than conventional training.
We also show that this \emph{Stochastic Weight Averaging} (SWA)
procedure finds much flatter solutions than SGD, and approximates the 
recent \emph{Fast Geometric Ensembling} (FGE) approach with a single model.
Using SWA we achieve notable improvement in test accuracy over
conventional SGD training on a range of state-of-the-art residual networks,
PyramidNets, DenseNets, and Shake-Shake networks on CIFAR-$10$, 
CIFAR-$100$, and ImageNet.  In short, SWA is extremely easy to implement, 
improves generalization, and has almost no computational overhead.
  
\end{abstract}

\section{INTRODUCTION}
\label{sec:intro}

With a better understanding of the loss surfaces for multilayer networks, we can accelerate the convergence, stability, and accuracy of training procedures in deep learning. Recent work \citep{garipov2018, draxler2018} shows 
that local optima found by SGD can be connected by simple curves of near constant loss.  Building upon this insight, \citet{garipov2018} also developed \emph{Fast Geometric Ensembling} (FGE) to sample multiple nearby points in weight space to create high performing ensembles in the time required to train a single DNN.

\begin{figure*}[!h]
	\centering
	\begin{subfigure}{0.27\textwidth}
		\includegraphics[width=\textwidth]{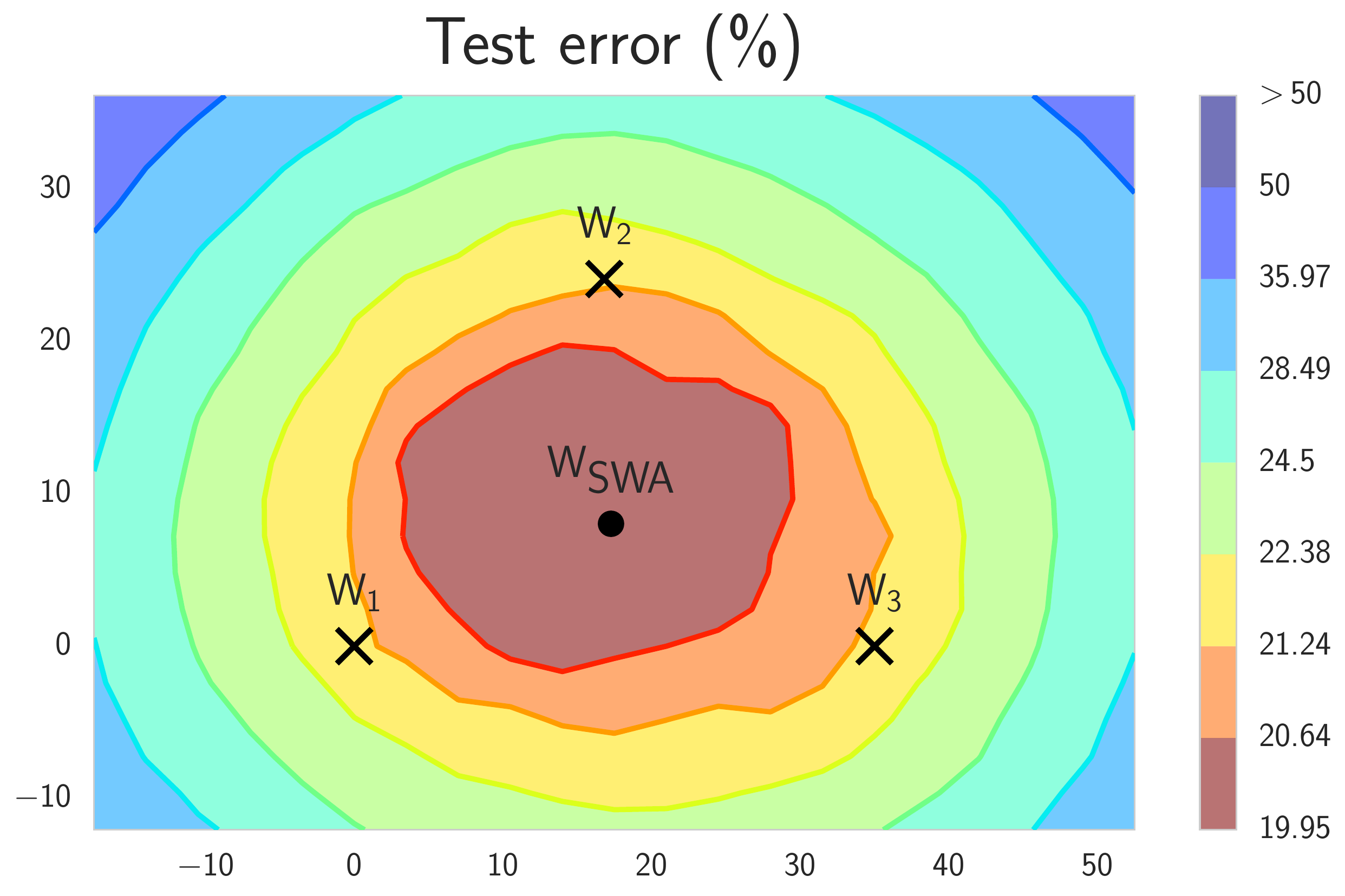}
	\end{subfigure}
	~~~~~~
	\begin{subfigure}{0.27\textwidth}
		\includegraphics[width=\textwidth]{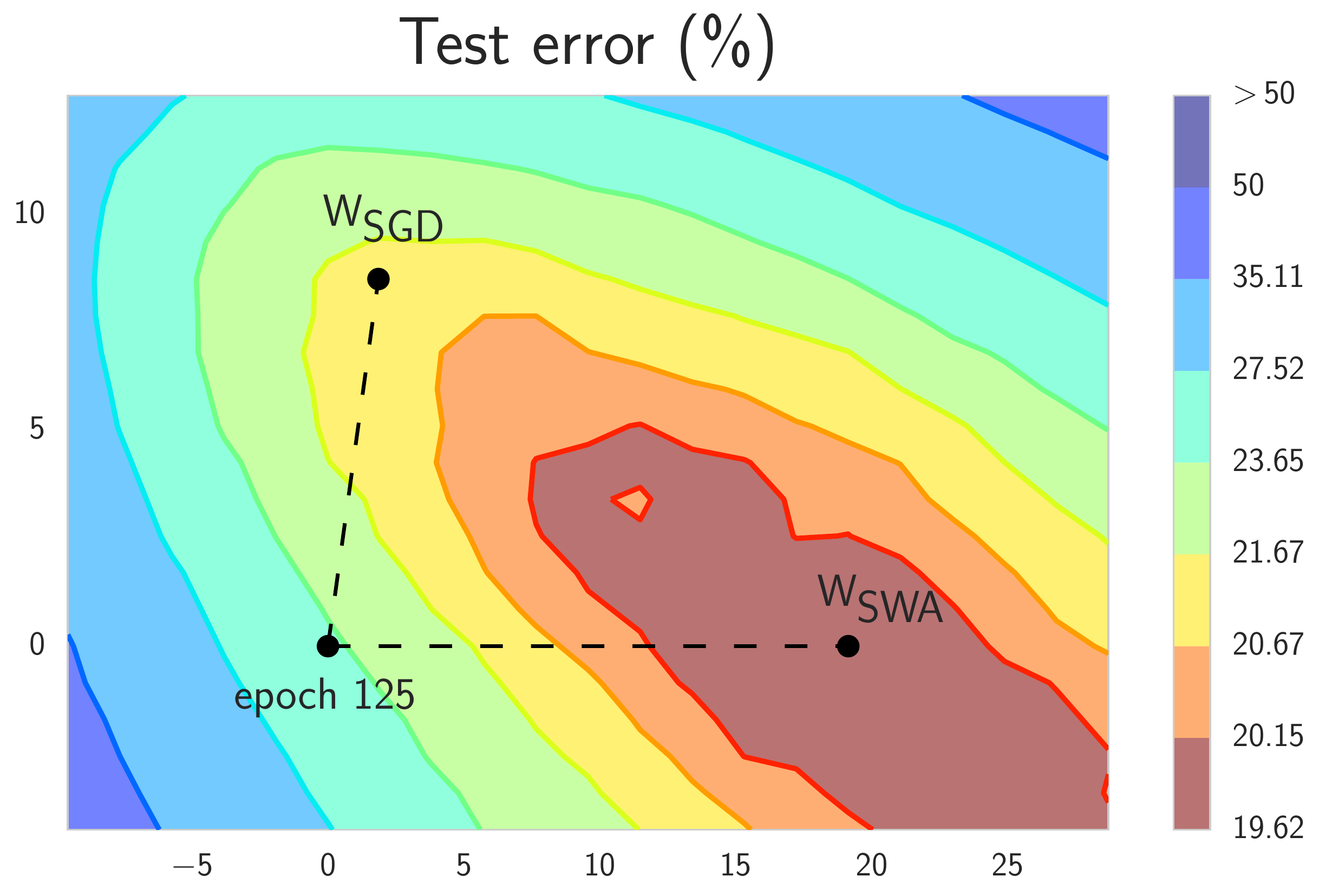}
	\end{subfigure}
	~~~~~~
	\begin{subfigure}{0.27\textwidth}
		\includegraphics[width=\textwidth]{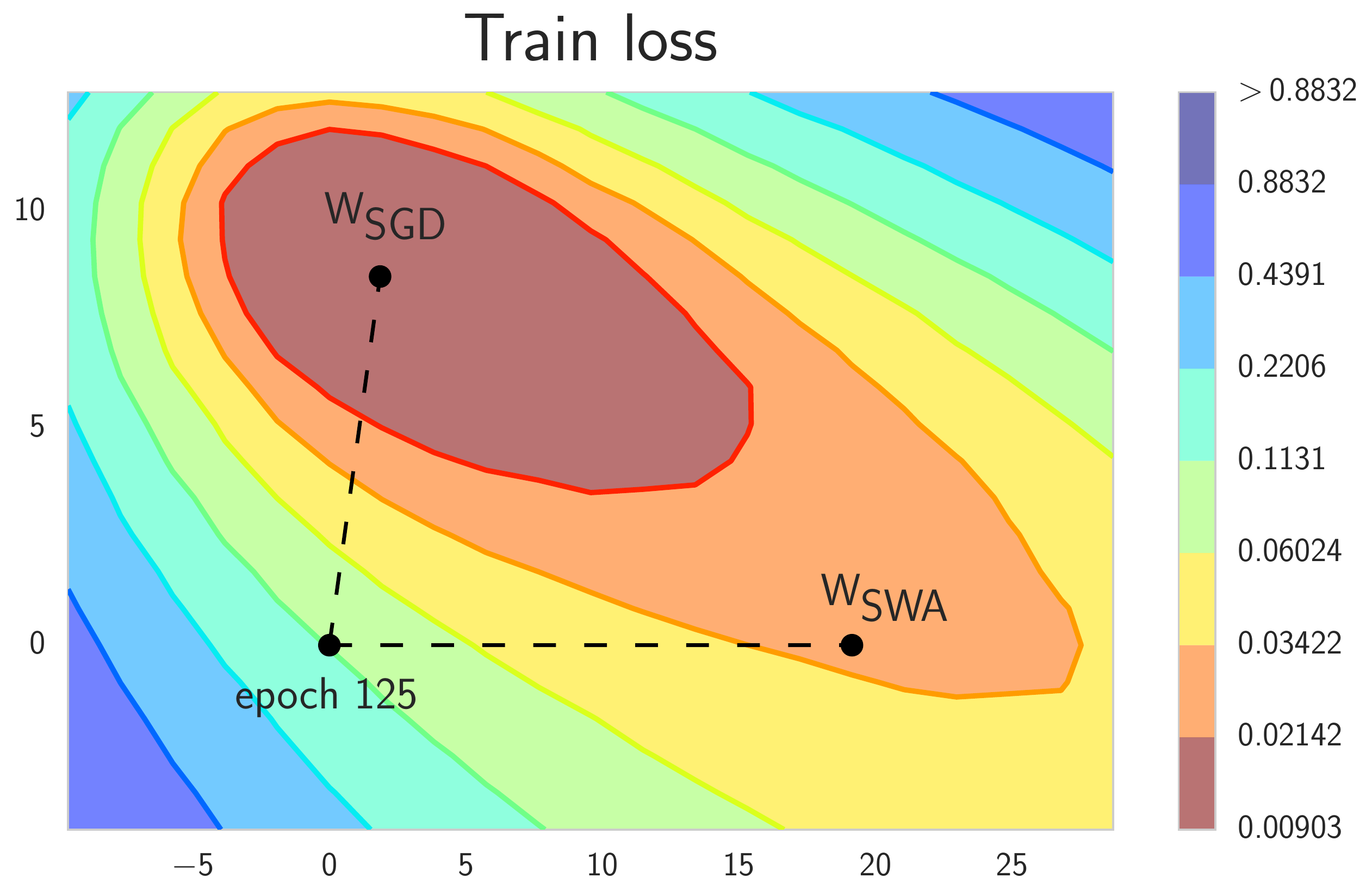}
	\end{subfigure}
  \caption{
    Illustrations of SWA and SGD with a Preactivation ResNet-$164$ on CIFAR-$100$\protect\footnotemark. 
    \textbf{Left}: test error surface for three
    FGE samples and the corresponding SWA solution (averaging in weight space). \textbf{Middle} and \textbf{Right}: 
    test error and train loss surfaces showing the weights proposed by SGD (at convergence)
    and SWA, starting from the same initialization of SGD after 125 training epochs.
	}
	\label{fig:intro}
\end{figure*}

FGE uses a high frequency cyclical learning rate with SGD to select networks to 
ensemble.  In Figure \ref{fig:intro} (left) we see that the weights of the networks 
ensembled by FGE are on the periphery of the most desirable solutions.
This observation suggests it is promising to  
average these points in \emph{weight space}, and use a network with 
these averaged weights, instead of forming an 
ensemble by averaging the outputs of networks in \emph{model space}.
Although the general idea of maintaining a running average of weights traversed by SGD dates back to \citet{ruppert1988}, 
this procedure is not typically used to train neural networks.  It is sometimes applied as an exponentially decaying running average in combination with a decaying learning rate (where it is called an exponential moving average), which smooths the trajectory of
conventional SGD but does not perform very differently. 
However, we show that an equally weighted average of the points traversed by SGD 
with a cyclical or high constant learning rate, which we refer to as \emph{Stochastic Weight Averaging} (SWA), has many surprising and
promising features for training deep neural networks, leading to a better understanding of the geometry of their loss surfaces.
Indeed, SWA with cyclical or constant learning rates can be used as a drop-in replacement for 
standard SGD training of multilayer networks --- but with improved generalization and essentially no overhead. In particular:

\begin{itemize}
  \item We show that SGD with cyclical
\citep[e.g.,][]{loshchilov2016} and
constant learning rates traverses regions of weight space corresponding 
to high-performing networks.  We find that while these models are moving 
around this optimal set they never reach its central points. We show that we 
can move into this more desirable space of points by averaging the weights proposed
over SGD iterations. 

\item While FGE ensembles \citep{garipov2018} can be trained in the same time as a 
single model, test predictions for an ensemble of $k$ models requires $k$ times more 
computation. We show that SWA can be interpreted as an approximation to FGE ensembles
but with the test-time, convenience, and interpretability of a single model.

\item We demonstrate that SWA leads to solutions that are wider than the optima
found by SGD.  \citet{keskar2017large} and \citet{hochreiter1997flat} conjecture
that the width of the optima is critically related to generalization. We
illustrate that the loss on the train is shifted with respect to the test
error (Figure \ref{fig:intro}, middle and right panels, and sections 
3, 4). We show that
SGD generally converges to a point near the boundary of the 
wide flat region of optimal points. SWA on the other hand is able to find a
point centered in this region, often with slightly worse train loss but with
substantially better test error. 

\item We show that the loss function is asymmetric in the direction connecting
SWA with SGD. In this direction, SGD is near the periphery of sharp ascent.
Part of the reason SWA improves generalization is that it finds solutions in flat regions 
of the training loss in such directions.

\item SWA achieves notable improvement for training a broad range of architectures 
over several consequential benchmarks. In particular, running SWA for just
  $10$ epochs on ImageNet we are able to achieve $0.8 \%$ improvement
  for ResNet-$50$ and DenseNet-$161$, and $0.6 \%$ improvement for
  ResNet-$150$. We achieve improvement of over $1.3\%$ on CIFAR-$100$ and
  of over $0.4\%$ on CIFAR-$10$ with Preactivation ResNet-$164$, VGG-$16$
  and Wide ResNet-$28$-$10$. We also achieve substantial improvement for
  the recent Shake-Shake Networks and PyramidNets. 

\item SWA is extremely easy to 
implement and has virtually no computational overhead compared to the conventional
training schemes. 

\item We provide an implementation of SWA at \\
\url{https://github.com/timgaripov/swa}.

\end{itemize}

We emphasize that SWA is finding a solution \emph{in the same basin of attraction} as SGD,
as can be seen in Figure~\ref{fig:intro}, but in a flatter region of the training loss. 
SGD typically 
finds points on the \emph{periphery} of a set of good weights. By running SGD with a cyclical
or high constant learning rate, we traverse the surface of this set of points, and by averaging
we find a more centred solution in a flatter region of the training loss. 
Further, the training loss for SWA is often slightly worse than 
for SGD suggesting that SWA solution is not a local optimum of the loss.
In the title of this paper, \emph{optima} is used in a general sense to mean \emph{solutions} (converged
points of a given procedure), rather than different local minima of the same objective. 

\section{RELATED WORK}
\label{sec:related_work}

This paper is fundamentally about better understanding
the geometry of loss surfaces and generalization in deep
learning. We follow the trajectory of weights traversed by 
SGD, leading to new geometric insights and the intuition 
that SWA will lead to better results than standard training.
Empirically, we make the discovery that SWA notably improves 
training of many state-of-the-art deep neural networks over a range
of consequential benchmarks, with essentially no overhead.

The procedures for training neural networks are constantly being improved.
New methods are being proposed for architecture design, regularization and 
optimization. The SWA approach is related to work in both optimization and
regularization. 
\footnotetext{
    Suppose we have three weight vectors $w_1, w_2, w_3$. 
    We set $u = (w_2 - w_1)$, $v = (w_3 - w_1) -\langle w_3 -~w_1, w_2 - ~w_1 \rangle /  \|w_2 - w_1\|^2 \cdot (w_2 - w_1)$.
    Then the normalized vectors $\hat u = u / \|u\|$, $\hat v = v / \|v\|$ form
    an orthonormal basis in the plane containing $w_1, w_2, w_3$. To visualize
    the loss in this plane, we define a Cartesian grid in the basis $\hat u, \hat v$
    and evaluate the networks corresponding to each of the points in the grid. A point
    $P$ with coordinates $(x, y)$ in the plane would then be given by 
    $P = w_1 + x \cdot \hat u + y \cdot \hat v$.
}

In optimization, there is great interest in how different types of local solutions 
affect generalization in deep learning. \citet{keskar2017large} claim that 
SGD is more likely to converge to broad local optima than batch gradient
methods, which tend to converge to sharp optima.  Moreover, they argue 
that the broad optima found by SGD are more likely to have good test 
performance, even if the training loss is worse than for the sharp optima.
On the other hand \citet{dinh2017} argue that all the known
definitions of sharpness are unsatisfactory and cannot on their own explain 
generalization.
\citet{chaudhari2016} propose the Entropy-SGD method that
explicitly forces optimization towards wide valleys. They report that although the 
optima found by Entropy-SGD are wider than those found by conventional SGD, 
the generalization performance is still comparable.

The SWA method is based on averaging multiple points along the trajectory
of SGD with cyclical or constant learning rates.  The general idea of maintaining
a running average of weights proposed by SGD was first considered in convex 
optimization by \citet{ruppert1988} and later by \citet{polyak1992}.  
However, this procedure is not typically used to train neural networks. Practitioners instead sometimes
use an exponentially decaying running average of the weights found by SGD with 
a decaying learning rate, which smooths the trajectory of SGD but performs comparably.

SWA is making use of multiple samples gathered through exploration of the 
set of points corresponding to high performing networks. To enforce exploration
we run SGD with constant or cyclical learning rates. \citet{mandt2017stochastic} 
show that under several simplifying assumptions running SGD with a constant
learning rate is equivalent to sampling from a Gaussian distribution centered
at the minimum of the loss, and the covariance of this Gaussian is controlled
by the learning rate. Following this explanation from \citep{mandt2017stochastic},
we can interpret points proposed by SGD as being constrained to the surface of a sphere, 
since they come from a high dimensional Gaussian distribution.  SWA effectively allows
us to go inside the sphere to find higher density solutions.

In a procedure called Fast Geometric Ensembling (FGE), \citet{garipov2018} showed that using a cyclical learning
rate it is possible to gather models that are spatially close to each other
but produce diverse predictions. They used the gathered models to train 
ensembles with no computational overhead compared to training a single 
DNN model.  In recent work \citet{neklyudov2018} also discuss an efficient approach
for model averaging of Bayesian neural networks.  SWA was inspired by 
following the trajectories of FGE proposals, in order to find a single model
that would approximate an FGE ensemble, but provide greater interpretability, convenience, and test-time 
scalability. 

Dropout \citep{srivastava2014dropout} is an extremely popular approach
to regularizing DNNs.  Across each mini-batch used for SGD, a 
different architecture is created by randomly dropping out neurons. The 
authors make analogies between dropout, ensembling, and Bayesian 
model averaging.  At test time, an ensemble approach is proposed, but
then approximated with similar results by multiplying each connection
by the dropout rate.  At a high level, SWA and Dropout are both at once
regularizers and training procedures, motivated to approximate an ensemble.
Each approach implements these high level ideas quite differently, and as we 
show in our experiments, can be combined for improved performance.

\section{STOCHASTIC WEIGHT AVERAGING}

We present Stochastic Weight Averaging (SWA) and analyze its
properties. In section \ref{sec:motivation}, we consider trajectories
of SGD with a constant and cyclical learning rate, which helps understand
the geometry of SGD training for neural networks, and motivates the SWA 
procedure. Then in section \ref{sec:method} we present the SWA algorithm
in detail, in section \ref{sec:complexity} we derive its complexity, and in section
\ref{sec:optima_width} we analyze the width of solutions found by 
SWA versus conventional SGD training. In section \ref{sec:ensembling} 
we then examine the relationship between SWA and the recently proposed
Fast Geometric Ensembling \citep{garipov2018}. Finally, in section
\ref{sec:geometry} we consider SWA from the perspective of 
stochastic convex optimization.

We note the name SWA has two meanings: on the one hand, it is an average of 
SGD weights.  On the other, with a cyclical or constant learning rate, SGD 
proposals are approximately sampling from the loss surface of the DNN, 
leading to stochastic weights.

\subsection{ANALYSIS OF SGD TRAJECTORIES}
\label{sec:motivation}

SWA is based on averaging the samples proposed by SGD using a learning rate
schedule that allows exploration of the region of weight space corresponding
to high-performing networks. In particular we
consider cyclical and constant learning rate schedules.

\begin{figure}[!t]
	\centering
	\includegraphics[width=0.45\textwidth]{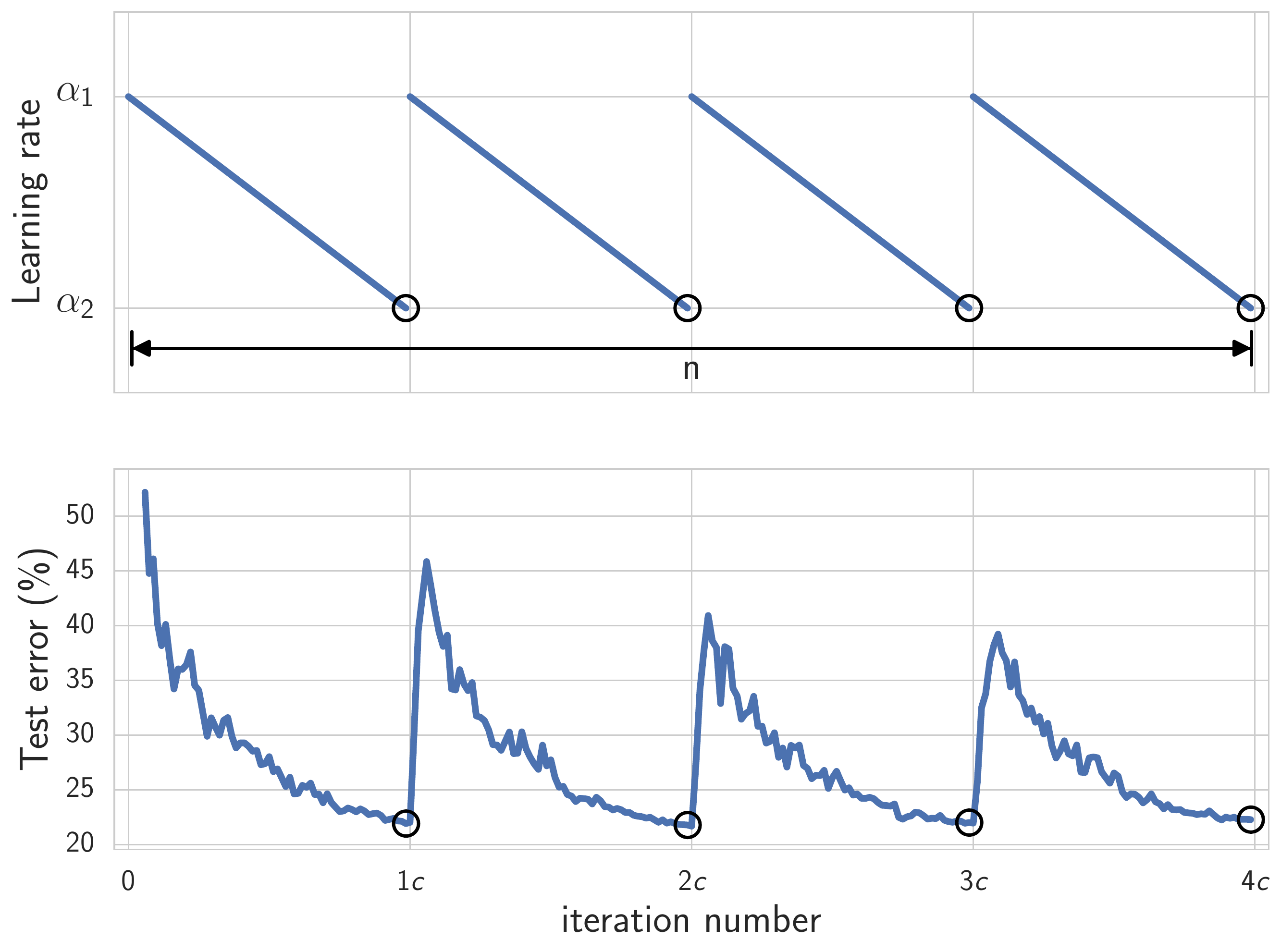}
	\caption{
        \textbf{Top}: cyclical learning rate as a function of iteration. 
        \textbf{Bottom}: test error as a function of iteration
        for cyclical learning rate schedule with Preactivation-ResNet-$164$ on 
        CIFAR-$100$.
        Circles indicate iterations corresponding to the minimum learning rates.
    }
	\label{fig:lr}    	
\end{figure}

The cyclical learning rate schedule that we adopt is inspired by 
\citet{garipov2018} and \citet{smith2017exploring}. 
In each cycle we linearly decrease the learning rate from $\alpha_1$ to
$\alpha_2$. The formula for the learning rate
at iteration $i$ is given by
\begin{align}
    \notag
    \alpha(i) &= 
		(1 - t(i)) \alpha_1 + t(i) \alpha_2,\\
    \notag
    t(i) &= \frac 1 c \left (\bmod (i - 1, c) + 1 \right).
\end{align}
The base learning rates $\alpha_1 \ge \alpha_2$ and the cycle length $c$  are the hyper-parameters of 
the method. 
Here by iteration we assume the processing of one batch of data.
Figure \ref{fig:lr} illustrates the 
cyclical learning rate schedule and the test error of the corresponding points.
Note that unlike the cyclical learning rate schedule of 
\citet{garipov2018} and \citet{smith2017exploring}, here we propose to use a 
discontinuous schedule that jumps directly from the minimum to maximum learning 
rates, and does \emph{not} steadily increase the learning rate as part of 
the cycle.  We use this more abrupt cycle because for our purposes exploration
is more important than the accuracy of individual proposals. For even greater 
exploration, we also consider constant learning rates $\alpha(i) = \alpha_1$.

\begin{figure*}[!h]
	\centering
	\begin{subfigure}{0.24\textwidth}
    \includegraphics[width=\textwidth]{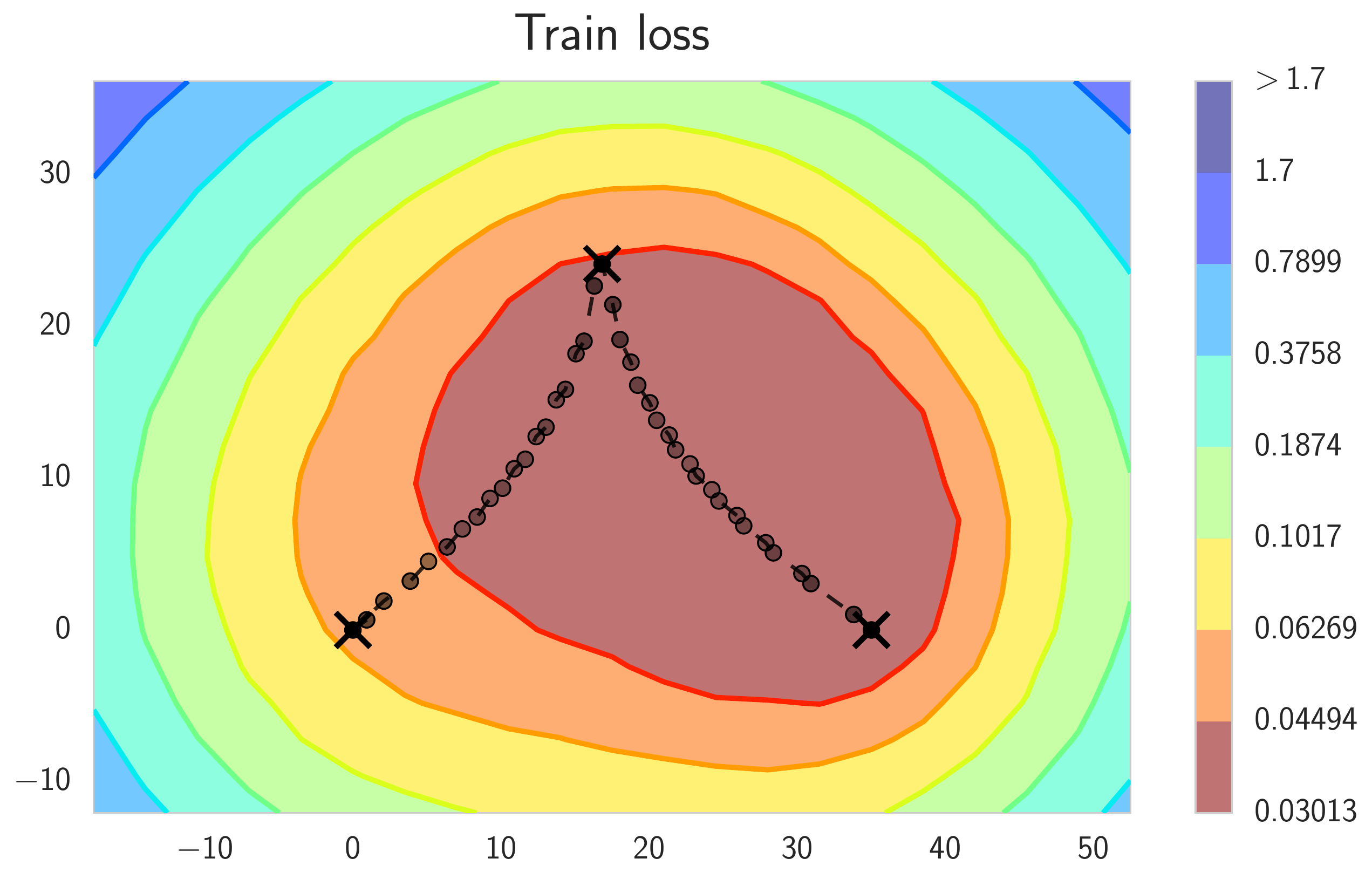}
	\end{subfigure}
	~
	\begin{subfigure}{0.24\textwidth}
		\includegraphics[width=\textwidth]{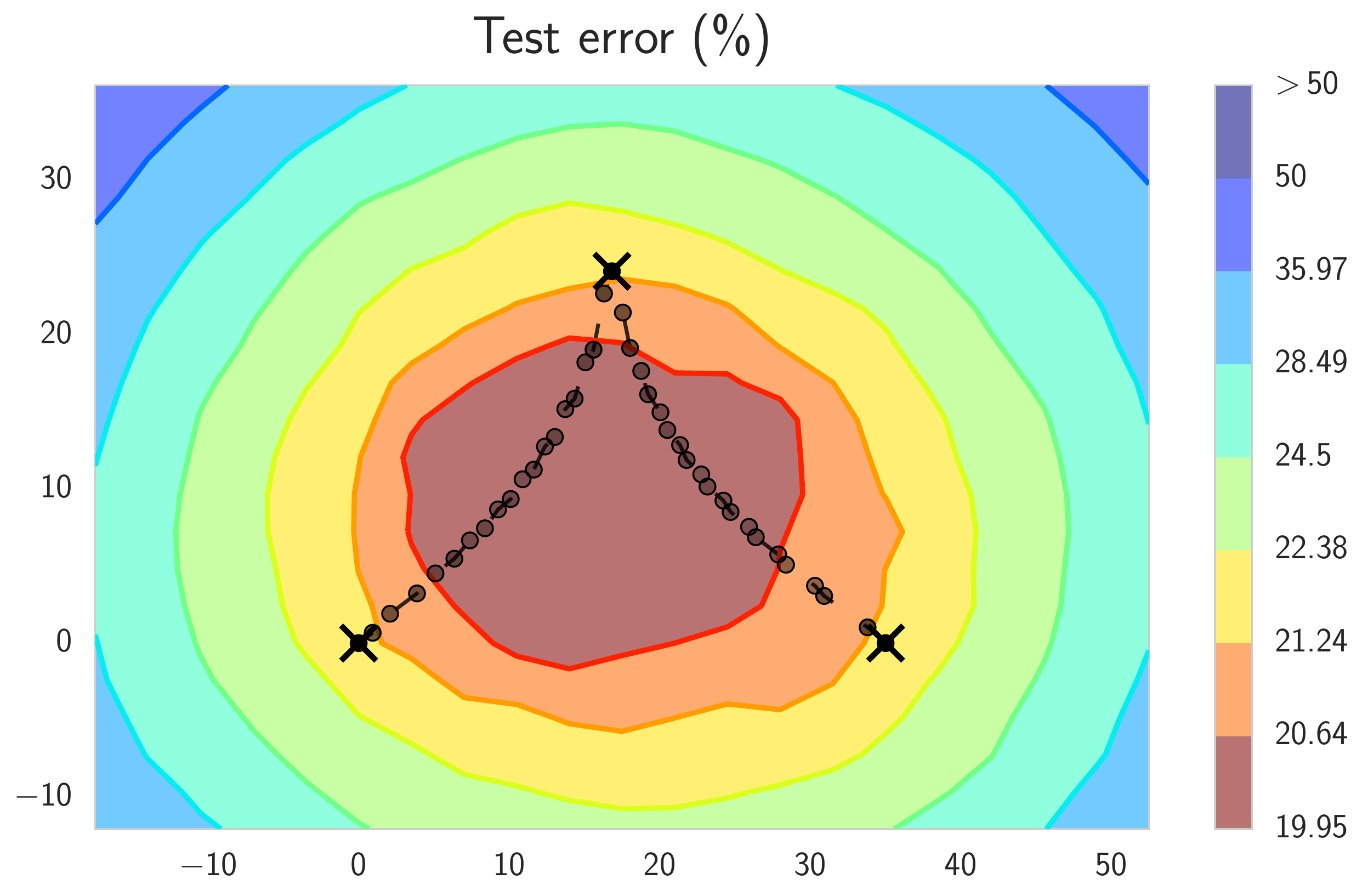}
	\end{subfigure}
	\begin{subfigure}{0.24\textwidth}
		\includegraphics[width=\textwidth]{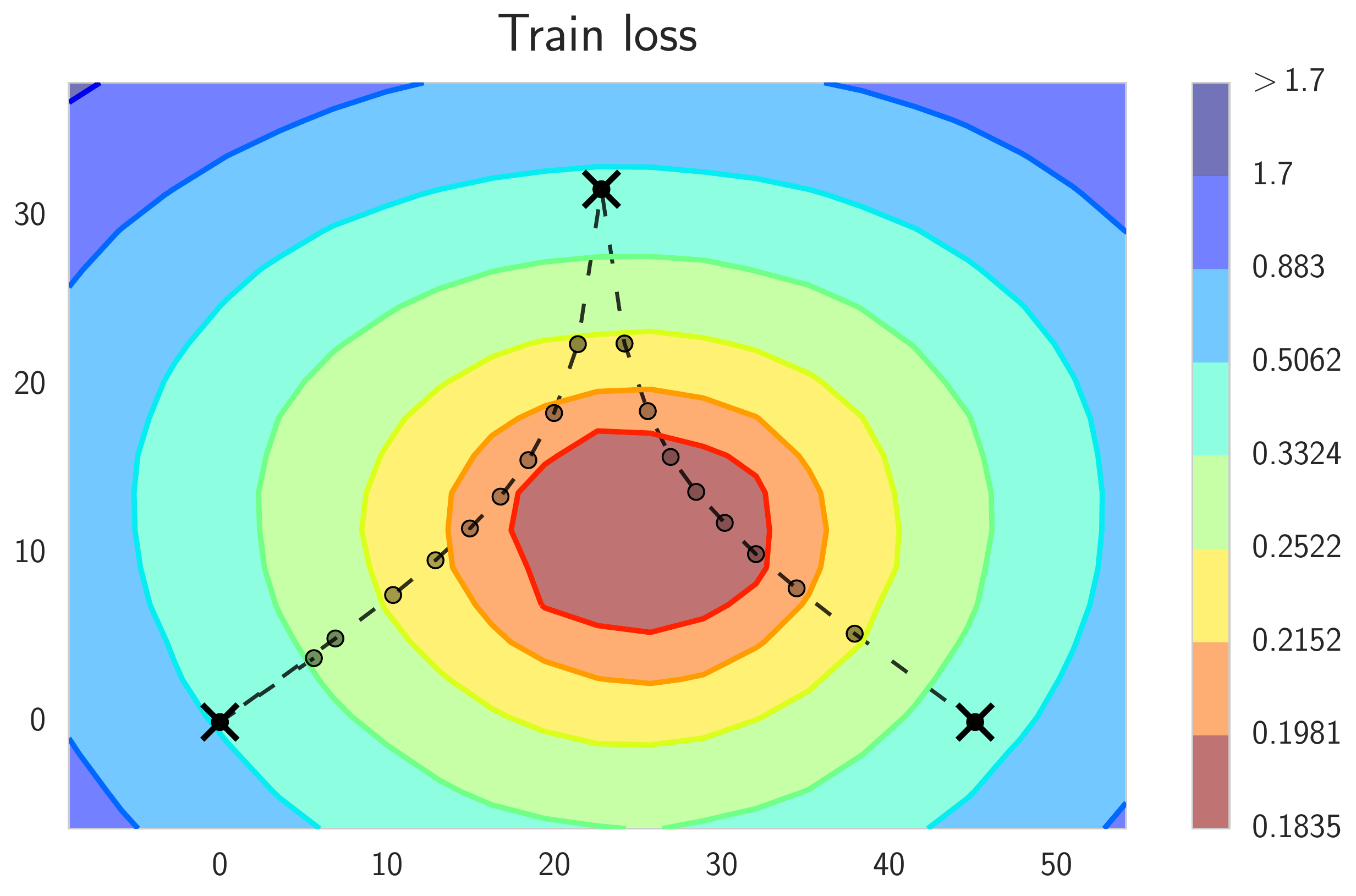}
	\end{subfigure}
	~
	\begin{subfigure}{0.24\textwidth}
		\includegraphics[width=\textwidth]{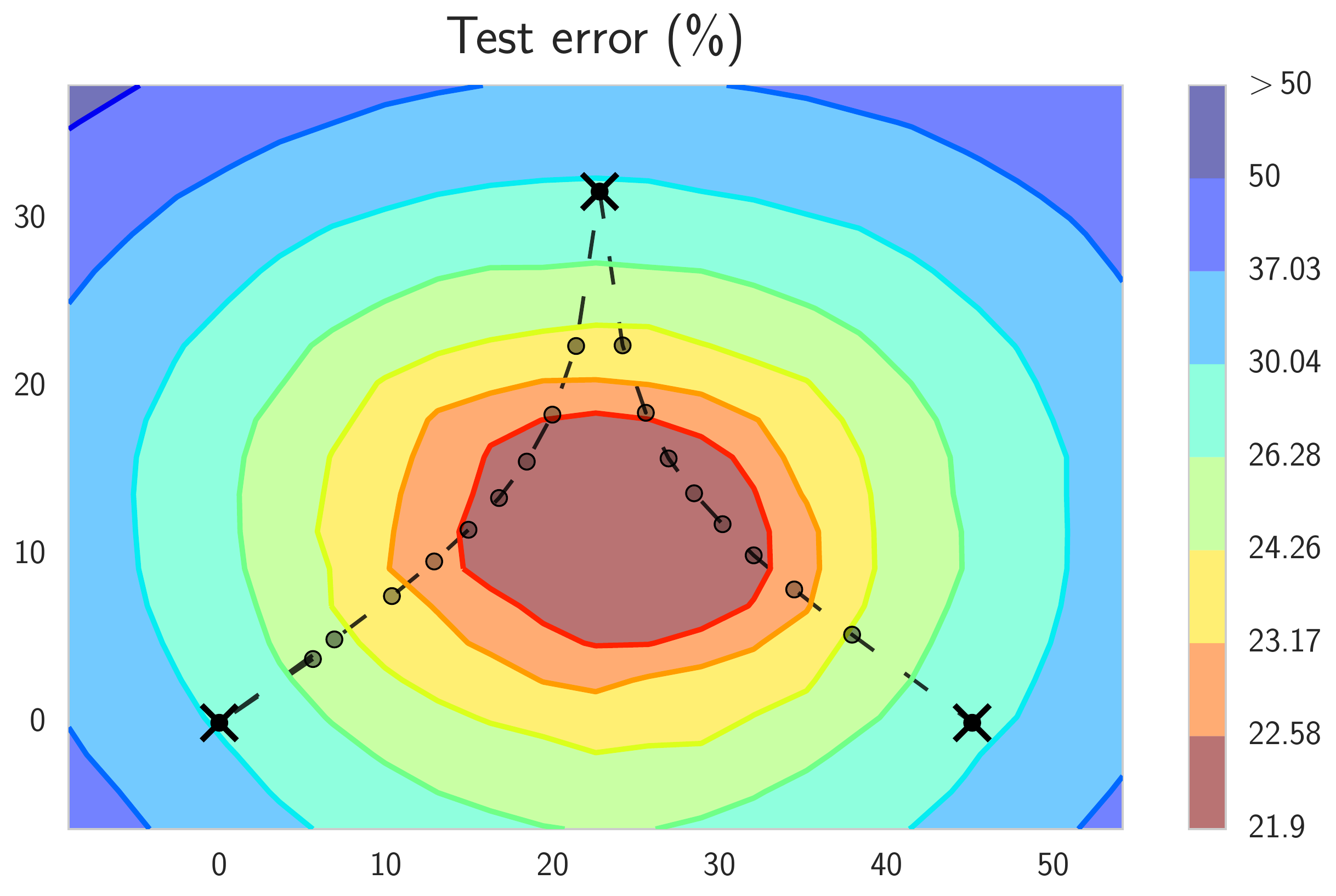}
	\end{subfigure}
	\caption{
      The $L_2$-regularized cross-entropy train loss and test error 
      surfaces of a Preactivation ResNet-$164$
	    on CIFAR-$100$ in the plane containing the first, middle and last points (indicated by black crosses)
      in the trajectories with (\textbf{left two}) cyclical 
      and (\textbf{right two}) constant learning rate schedules.
  }
	\label{fig:loss_error_visualizations}
\end{figure*}

We run SGD with cyclical and constant learning rate schedules starting
from a pretrained point for a Preactivation ResNet-$164$ on CIFAR-$100$. 
We then use the first, middle and last point of each of the trajectories to define
a $2$-dimensional plane in the weight space containing all affine combinations
of these points. 
In Figure \ref{fig:loss_error_visualizations}
we plot the loss on train and error on test for points in these planes.
We then project the other points of the trajectory to the plane of the plot.
Note that the trajectories do not generally lie in the plane of the plot,
except for the first, last and middle points, showed by black crosses in the
figure. Therefore for other points of the trajectories it is not possible to tell the
value of train loss and test error from the plots.

The key insight from Figure \ref{fig:loss_error_visualizations} is that both
methods explore points close to the periphery of the set of high-performing
networks.
The visualizations suggest that both methods are doing exploration in the
region of space corresponding to DNNs with high accuracy. The main difference
between the two approaches is that the individual proposals of SGD with a 
cyclical learning rate schedule are in general much more accurate than the proposals
of a fixed-learning rate SGD.  After making a large step,
SGD with a cyclical learning rate spends several epochs fine-tuning the 
resulting point with a decreasing learning rate. SGD with a fixed learning rate on the other hand is always
making steps of relatively large sizes, exploring more efficiently than with 
a cyclical learning rate, but the individual proposals are worse.

Another important insight we can get from Figure \ref{fig:loss_error_visualizations}
is that while the train loss and test error surfaces are qualitatively similar,
they are not perfectly aligned. The shift between train and test suggests that
more robust central points in the set of high-performing networks
can lead to better generalization.
Indeed, if we average several proposals from the optimization trajectories, we get
a more robust point that has a substantially higher test performance than the 
individual proposals of SGD, and is essentially centered on the shifted mode for
test error. We further discuss the reasons
for this behaviour in sections \ref{sec:optima_width}, \ref{sec:ensembling},
\ref{sec:geometry}. 

\raggedbottom

\subsection{SWA ALGORITHM}
\label{sec:method}

We now present the details of the Stochastic Weight Averaging algorithm, 
a simple but effective modification for training neural networks, motivated by
our observations in section \ref{sec:motivation}.

Following \citet{garipov2018}, we start with a pretrained model $\hat w$. 
We will refer to the number of epochs required to train a given DNN
with the conventional training procedure as its training budget and will
denote it by $B$.
The pretrained model $\hat w$ can be trained with the conventional training 
procedure for full training budget or reduced number of epochs (e.g. $0.75 B$).
In the latter case we just stop the training early without modifying the 
learning rate schedule.
Starting from $\hat w$ we continue
training, using a cyclical or constant learning rate schedule. 
When using a cyclical learning rate we capture the 
models $w_i$ that correspond to the minimum values of the learning rate 
(see Figure \ref{fig:lr}), following \citet{garipov2018}. For constant 
learning rates we capture models at each epoch.
Next, we average the weights of all the captured networks $w_i$
to get our final model $w_{\text{SWA}}$. 

Note that for cyclical learning rate schedule, the SWA algorithm is 
related to FGE \citep{garipov2018}, except that instead of
averaging the predictions of the models, we average their weights,
and we use a different type of learning rate cycle.
In section \ref{sec:ensembling} we show how SWA can approximate
FGE, but with a single model.

\paragraph{Batch normalization.} If the DNN uses batch normalization \citep{ioffe2015}, we 
run one additional pass over the data, as in \citet{garipov2018}, to compute the running mean 
and standard deviation of 
the activations for each layer of the network with $w_{\text{SWA}}$ weights after the training is 
finished, since these statistics are not collected during training. For most deep learning libraries,
such as PyTorch or Tensorflow, one can typically collect these statistics by making a forward pass 
over the data in training mode.

The SWA procedure is summarized in Algorithm \ref{alg:SWA}.

\begin{algorithm}[!t]
    \caption{Stochastic Weight Averaging}
    \label{alg:SWA}
\begin{algorithmic}
    \REQUIRE ~\\ weights $\hat w$, LR bounds $\alpha_1, \alpha_2$,\\ 
    cycle length~$c$~(for constant learning rate $c=1$), number of iterations $n$
    \ENSURE $w_{\text{SWA}}$ 
    \STATE $w \leftarrow \hat w$ \COMMENT{Initialize weights with $\hat w$}
    \STATE $w_{\text{SWA}} \leftarrow w$ 
    \FOR {$i \leftarrow 1, 2, \ldots, n$}
        \STATE $\alpha \leftarrow \alpha(i)$ \COMMENT{Calculate LR for the iteration}
        \STATE $w \leftarrow w - \alpha \nabla \mathcal{L}_i(w)$ \COMMENT{Stochastic gradient update}
        \IF{$\bmod(i, c) = 0$} 
            \STATE {$n_{\text{models}} \leftarrow i / c$} \COMMENT{Number of models}
            \STATE {$w_{\text{SWA}} \leftarrow \frac{w_{\text{SWA}} \cdot n_{\text{models}} + w}{n_{\text{models}} + 1}$} \COMMENT{Update average}
        \ENDIF
    \ENDFOR\\
    \{Compute BatchNorm statistics for $w_{\text{SWA}}$ weights\}
\end{algorithmic}
\end{algorithm}

\subsection{COMPUTATIONAL COMPLEXITY}
\label{sec:complexity}

The time and memory overhead of SWA compared to conventional training is negligible.
During training, we need to maintain a copy of the running average of DNN weights.
Note however that the memory consumption in storing a DNN 
is dominated by its activations rather than its weights, and thus is 
only slightly increased by the SWA procedure, even for large DNNs
(e.g., on the order of 10\%).
After the training is complete 
we only need to store the model that aggregates the average,
 leading to the same memory requirements
as standard training.

During training extra time is only spent to update the aggregated weight average.
This operation is of the form
\begin{align}
  \notag
  w_{\text{SWA}} \leftarrow \frac{w_{\text{SWA}} \cdot n_{\text{models}} + w}{n_{\text{models}} + 1} \,,
\end{align}
and it only requires computing a weighted sum of the weights of two DNNs. As we
apply this operation at most once per epoch, SWA and SGD require practically the same 
amount of computation. Indeed, a similar operation is performed as a part of each 
gradient step, and each epoch consists of hundreds of gradient steps. 

\subsection{SOLUTION WIDTH}
\begin{figure*}[!h]
	\centering
	\begin{subfigure}{0.42\textwidth}
		\includegraphics[width=\textwidth]{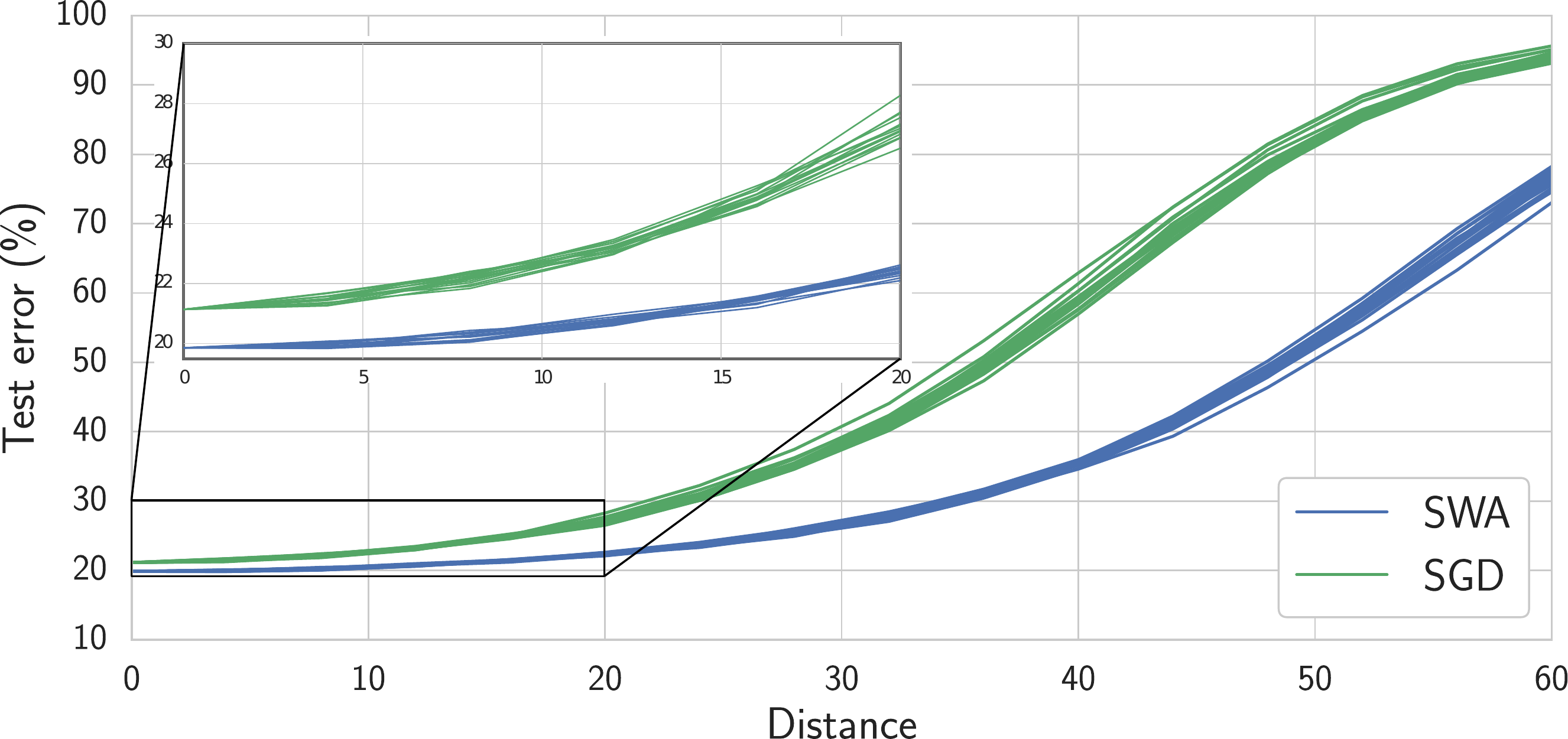}
	\end{subfigure}
	~~~~
	\begin{subfigure}{0.42\textwidth}
		\includegraphics[width=\textwidth]{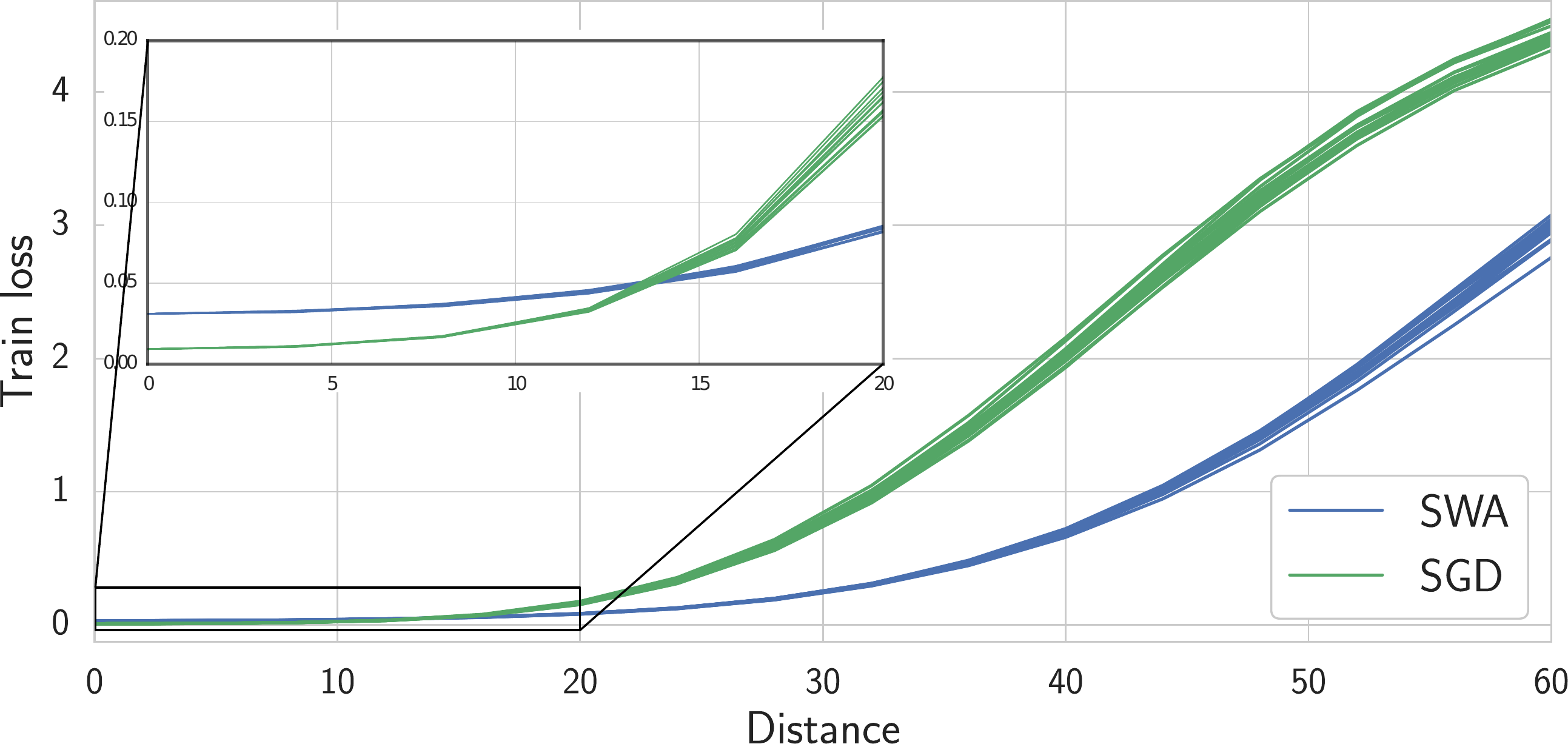}
	\end{subfigure}
	\caption{
      (\textbf{Left}) Test error and (\textbf{Right}) $L_2$-regularized 
      cross-entropy train loss as a function of 
      a point on a random ray starting at SWA (blue) and SGD (green) solutions
      for Preactivation ResNet-$164$ on CIFAR-$100$.
      Each line corresponds to a different random ray.
  }
	\label{fig:rand_ray}
\end{figure*}
\label{sec:optima_width}
\begin{figure*}[!h]
	\centering
	\centering
	\begin{subfigure}{0.42\textwidth}
	  \includegraphics[width=\textwidth]{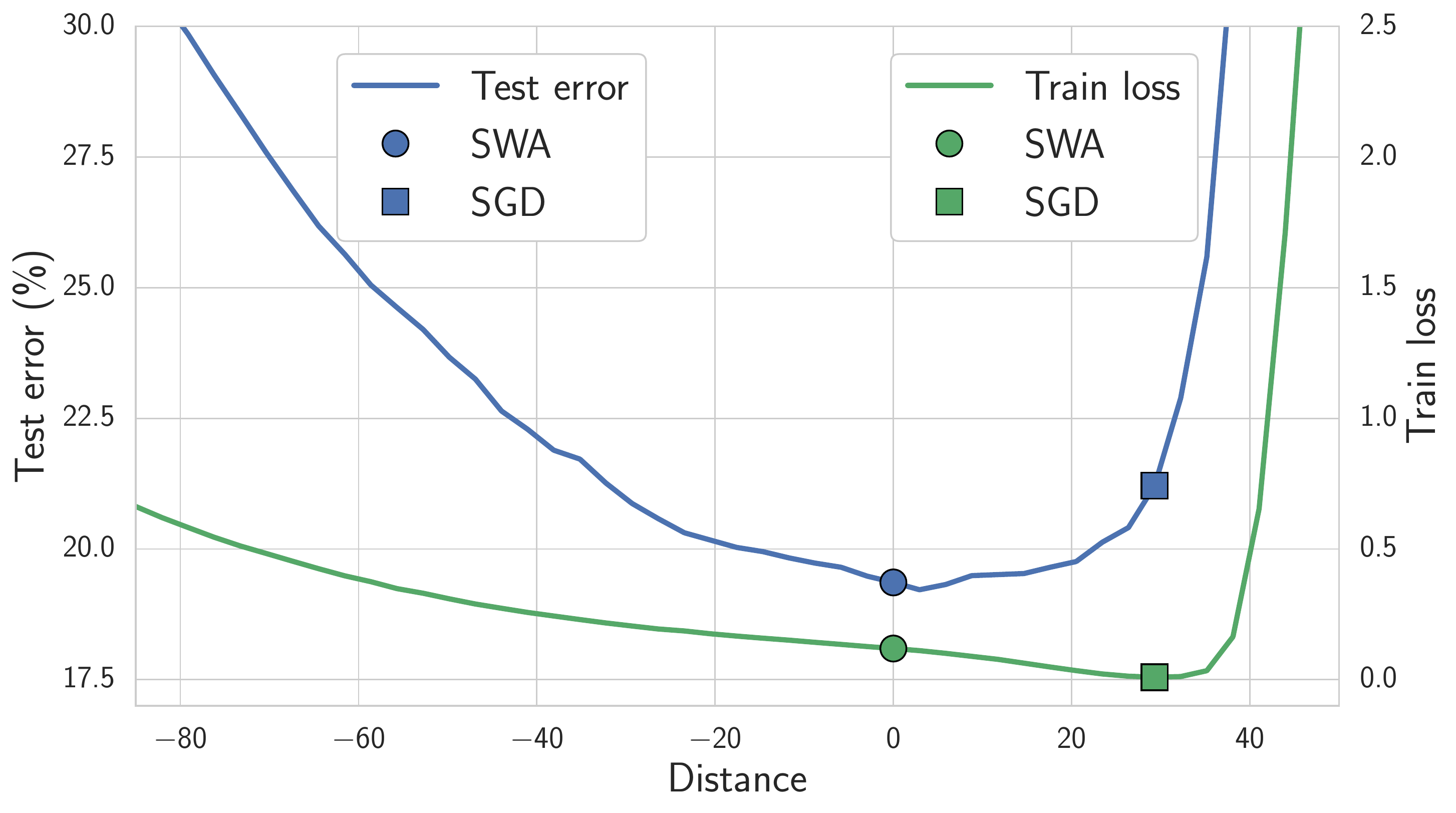}
	\end{subfigure}
	~~~~
	\begin{subfigure}{0.42\textwidth}
	  \includegraphics[width=\textwidth]{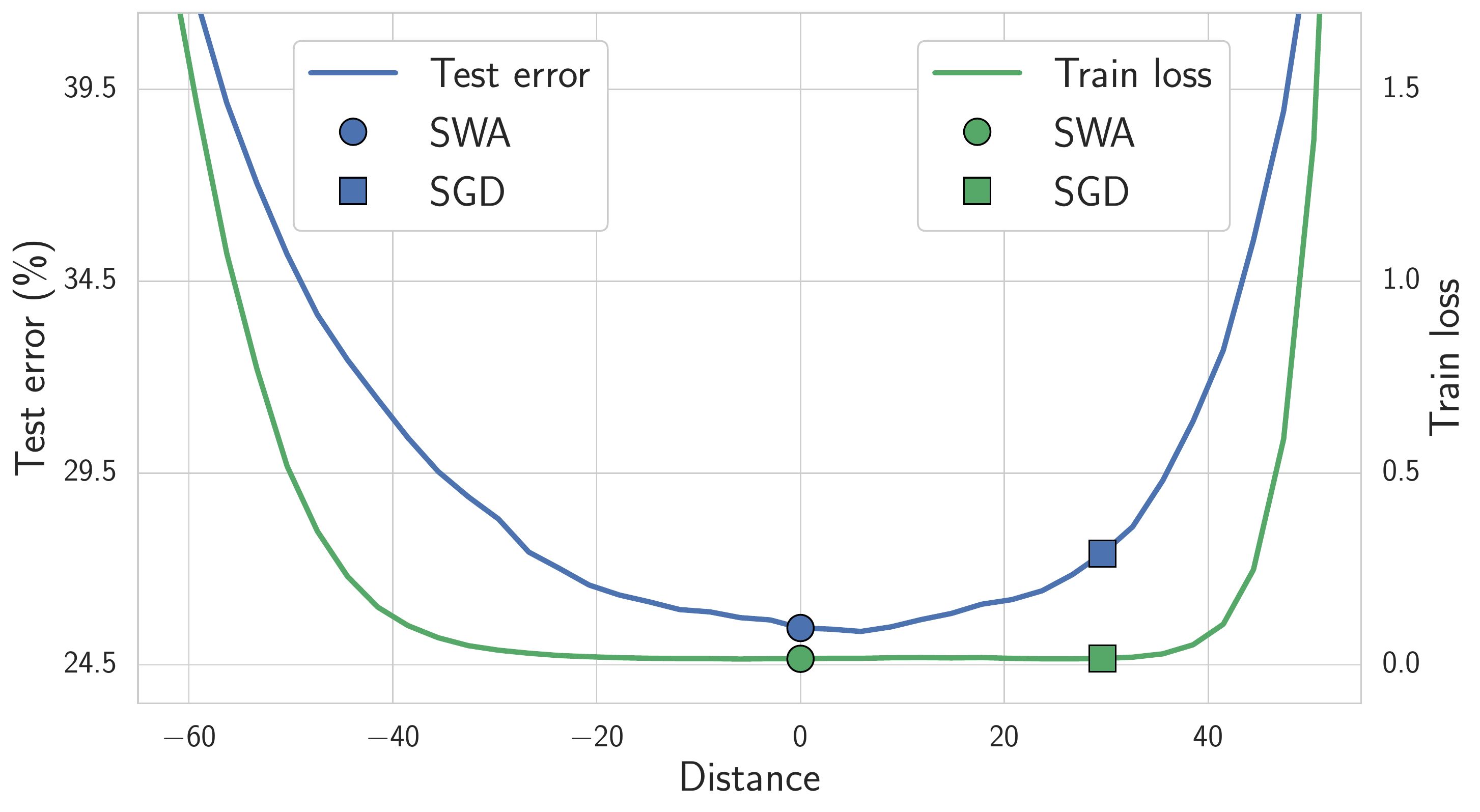}
	\end{subfigure}
	\caption{
      $L_2$-regularized cross-entropy train loss and test error as a function of
      a point on the line connecting SWA and SGD solutions on CIFAR-$100$.
      \textbf{Left}: Preactivation ResNet-$164$. \textbf{Right}: VGG-$16$.
      }
	\label{fig:optima_seg}
\end{figure*}

\citet{keskar2017large} and \citet{chaudhari2016} conjecture that the width of a local
optimum is related to generalization.
The general explanation for the importance of width is that the surfaces of train
loss and test error are shifted with respect to each other and it is thus desirable
to converge to the modes of broad optima, which stay approximately optimal under small 
perturbations. In this section we compare the solutions found by SWA and SGD
and show that SWA generally leads to much wider solutions.

Let $w_{\text{SWA}}$ and $w_{\text{SGD}}$ denote the weights of DNNs trained 
using SWA and conventional SGD, respectively. 
Consider the rays 
\begin{align}
  \notag
  w_{\text{SWA}}(t, d) = w_{\text{SWA}} + t \cdot d,\\
  \notag
  w_{\text{SGD}}(t, d) = w_{\text{SGD}} + t \cdot d,
\end{align}
which follow a direction vector $d$ on the unit sphere,
starting at $w_{\text{SWA}}$ and $w_{\text{SGD}}$, respectively. In Figure~\ref{fig:rand_ray} 
we plot train loss and test error of $w_{\text{SWA}}(t, d_i)$ and
$w_{\text{SGD}}(t, d_i)$ as a function of $t$ for $10$ random directions $d_i$,
$i = 1, 2, \ldots, 10$ drawn from a uniform distribution on the unit sphere. 
For this visualization we use a Preactivation ResNet-$164$ on CIFAR-$100$.

First, while the loss values on train for $w_{\text{SGD}}$ and $w_{\text{SWA}}$
are quite similar (and in fact $w_{\text{SGD}}$ has a slightly lower train loss),
the test error for $w_{\text{SGD}}$ is lower by $1.5 \%$ (at the converged value
corresponding to $t=0$).
Further, the shapes of both train loss and test error curves
are considerably wider for $w_{\text{SWA}}$ than for $w_{\text{SGD}}$, suggesting
that SWA indeed converges to a wider solution: we have to step much further away
from $w_{\text{SWA}}$ to increase error by a given amount. 
We even see the error curve for SGD has an inflection point that is not present for 
these distances with SWA.

Notice that in Figure \ref{fig:rand_ray} any of the random directions from $w_{\text{SGD}}$
increase test error.
However, we know that the direction from $w_{\text{SGD}}$ to $w_{\text{SWA}}$ would 
decrease test error,
since $w_{\text{SWA}}$ has considerably lower test error than $w_{\text{SGD}}$.
In other words, the path from $w_{\text{SGD}}$ to $w_{\text{SWA}}$ is qualitatively 
different from all directions shown in Figure \ref{fig:rand_ray}, because
along this direction $w_{\text{SGD}}$ is far from optimal.
We therefore consider the line segment connecting 
 $w_{\text{SGD}}$ and $w_{\text{SWA}}$:
\begin{align}
  \notag
  w(t) = t \cdot w_{\text{SGD}} + (1 - t) \cdot w_{\text{SWA}} \,.
\end{align}
In Figure \ref{fig:optima_seg} we plot the train loss
and test error of $w(t)$ as a function of signed distance from $w_{\text{SWA}}$ for 
Preactivation ResNet-$164$ and VGG-$16$ on CIFAR-$100$. 

We can extract several key insights about $w_{\text{SWA}}$ and $w_{\text{SGD}}$ from 
Figure \ref{fig:optima_seg}. First, the train loss and test error plots are 
indeed substantially shifted, and the point obtained by minimizing the train
loss is far from optimal on test. Second, $w_{\text{SGD}}$ lies near the boundary
of a wide flat region of the train loss. Further, the loss is very steep near
$w_{\text{SGD}}$. 

\citet{keskar2017large} argue that the loss near sharp optima found by SGD with very
large batches are actually flat in most directions, but there
exist directions in which the optima are extremely steep. They conjecture that
because of this sharpness the generalization performance of large batch optimization
is substantially worse than that of solutions found by small batch SGD.
Remarkably, in our experiments in this section we observe that there exist directions of steep 
ascent even for small batch optima, and that SWA provides even wider solutions 
(at least along random directions) with better generalization. Indeed, we can see clearly in 
Figure~\ref{fig:optima_seg} that SWA is not finding a different minima than SGD, but rather
a flatter region in the same basin of attraction. We can also see clearly that the significant 
asymmetry of the loss function in certain directions, such as the direction SWA to SGD, has a role in 
understanding why SWA provides better generalization than SGD. In these directions SWA
finds a much flatter solution than SGD, which can be near the periphery of sharp ascent.

\subsection{CONNECTION TO ENSEMBLING}
\label{sec:ensembling}

\citet{garipov2018} proposed the Fast Geometric Ensembling (FGE) procedure 
for training ensembles in the time required to train a single model. Using a
cyclical learning rate, FGE generates a sequence of points
that are close to each other in the weight space, but produce diverse 
predictions. In SWA instead of averaging the predictions of the models we 
average their weights. However, the predictions proposed by FGE ensembles
and SWA models have similar properties. 

Let $f(\cdot)$ denote the predictions of a neural network parametrized by
weights $w$.
We will
assume that $f$ is a scalar (e.g.\ the probability for a particular class)
twice continuously differentiable function with respect to $w$. 

Consider points $w_i$ proposed by FGE. These points are close in the weight space by 
design, and 
concentrated around their 
average $w_{\text{SWA}} = \frac 1 n \sum_{i=1}^n w_i$. We denote
$\Delta_i = w_i - w_{\text{SWA}}$. Note $\sum_{i=1}^n \Delta_i = 0$.
Ensembling the networks corresponds to averaging
the function values
\begin{align}
  \notag
  \bar{f} = \frac 1 n \sum_{i=1}^n f(w_i).
\end{align}
Consider the linearization of $f$ at $w_{\text{SWA}}$.
\begin{align}
  \notag
  f(w_j) = f(w_{\text{SWA}}) +
  \left \langle \nabla f(w_{\text{SWA}}), \Delta_j \right\rangle + 
  O(\|\Delta_j\|^2),
\end{align}
where $\langle \cdot, \cdot \rangle$ denotes the dot product.
Thus, the difference between averaging the weights and averaging the
predictions
\begin{align}
  \notag
  \bar{f} - f(w_{\text{SWA}}) = 
  \frac{1}{n} \sum_{i=1}^n 
  \left( \left \langle \nabla f(w_{\text{SWA}}),  \Delta_i \right\rangle + 
  O(\|\Delta_i\|^2)\right) \\ 
  = \left \langle \nabla f(w_{\text{SWA}}), \frac{1}{n} \sum_{i=1}^n \Delta_i \right\rangle + 
  O(\Delta^2)
  = O(\Delta^2), \notag
\end{align}
where $\Delta = \max_{i=1}^n \|\Delta_i\|$.
Note that the difference between the predictions of different perturbed networks
is
\begin{align}
  \notag
  f(w_i) - f(w_j) = 
  \langle \nabla f(w_{\text{SWA}}), \Delta_i - \Delta_j \rangle + O(\Delta^2),
\end{align}
and is thus of the first order of smallness, while the difference between
averaging predictions and averaging weights is of the second order of
smallness. Note that for the points proposed by FGE the distances
between proposals are relatively small by design, which justifies 
the local analysis.

To analyze the difference between ensembling and averaging the weights of
FGE proposals in practice, we run FGE for $20$ epochs and compare the
predictions of different models on the test dataset with a Preactivation
ResNet-164 \citep{he2016deep} on CIFAR-$100$. The norm of the difference
between the class probabilities of consecutive FGE proposals averaged over 
the test 
dataset is $0.126$. We then average the weights of the proposals and compute
the class probabilities on the test dataset. The norm of difference of the probabilities for
the SWA model and the FGE ensemble is $0.079$, which is
substantially smaller than the difference between the probabilities of consecutive
FGE proposals. Further, the fraction of objects for which consecutive FGE proposals 
output the same labels is not greater than $87.33 \%$. For FGE and SWA
the fraction of identically labeled objects is $95.26 \%$.

The theoretical considerations and empirical results presented in this section
suggest that SWA can approximate the FGE ensemble
with a single model.

\subsection{CONNECTION TO CONVEX MINIMIZATION}
\label{sec:geometry}
\begin{table*}[!th]
	\caption{Accuracies ($\%$) of SWA, SGD and FGE methods on CIFAR-100 and CIFAR-10 datasets for different training budgets. Accuracies for the FGE ensemble are from \citet{garipov2018}.}
	\label{table:main_experiments}
	\centering
	\begin{tabular}{cccccc}
		\toprule
	 & & & \multicolumn{3}{c}{SWA}	\\
	 \cline{4-6}
	 \\[-1em]	
    DNN (Budget) & SGD & FGE ($1$ Budget) & $1$ Budget & $1.25$ Budgets & $1.5$ Budgets \\
		\hline
		\multicolumn{6}{c}{\small CIFAR-100}\\
		\hline
    VGG-16 ($200$) &
    $72.55 \pm 0.10$ & $74.26$ & $73.91 \pm 0.12$ & $74.17 \pm 0.15$ & $74.27 \pm 0.25$ \\
    ResNet-164 ($150$) &
    $78.49 \pm 0.36$ & $79.84$ & $79.77 \pm 0.17$ & $80.18 \pm 0.23$ & $80.35 \pm 0.16$ \\
    WRN-28-10 ($200$) &
    $80.82 \pm 0.23$ & $82.27$ & $81.46 \pm 0.23$ & $81.91 \pm 0.27$ & $82.15 \pm 0.27$ \\
    PyramidNet-$272$ ($300$) &
    $ 83.41 \pm 0.21 $ & --& -- & $83.93 \pm 0.18 $ & $84.16 \pm 0.15$\\
		\hline 
    \multicolumn{6}{c}{\small CIFAR-10}\\
		\hline 
    VGG-16 ($200$) &
    $93.25 \pm 0.16$ & $93.52$ & $93.59 \pm 0.16$ & $93.70 \pm 0.22$ & $93.64 \pm 0.18$ \\
    ResNet-164 ($150$) &
    $95.28 \pm 0.10$ & $95.45$ & $95.56 \pm 0.11$ & $95.77 \pm 0.04$ & $95.83 \pm 0.03$ \\
    WRN-28-10 ($200$) &
    $96.18 \pm 0.11$ &$96.36$& $96.45 \pm 0.11$ & $96.64 \pm 0.08$ & $96.79 \pm 0.05$ \\
    ShakeShake-2x64d ($1800$) &
    $96.93 \pm 0.10$ & -- & -- & $97.16 \pm 0.10$ & $97.12 \pm 0.06$ \\
		\bottomrule   
	\end{tabular}
\end{table*}

\citet{mandt2017stochastic} showed that under strong simplifying assumptions SGD with 
a fixed learning rate approximately samples from a Gaussian distribution 
centered at the minimum of the loss. Suppose this is the case when we run
SGD with a fixed learning rate for training a DNN. 

Let us denote the dimensionality of the weight space of the neural network by $d$.
Denote the samples produced by SGD by $w_{i},~i=1, 2, \ldots, k$. Assume the points
$w_i$ are concentrated around the local optimum $\hat w$. The SWA
solution is given by $w_{\text{SWA}} = \frac 1 n \sum_{i=1}^k w_i$.
The points $w_i$ are samples from a multidimensional Gaussian $\mathcal{N}(\hat w, \Sigma)$
for some covariance matrix $\Sigma$ defined by the curvature of the loss, batch
size and the learning rate. Note that the samples from a multidimensional 
Gaussian are concentrated on the ellipsoid
\begin{align}
  \notag
  \left\{ z \in \mathbb{R}^d \vert~~ \|\Sigma^{-\frac{1}{2}} (z - \hat w)\| = \sqrt d\right\},
\end{align}
and the probability mass for a sample to end up inside the ellipsoid near 
$\hat w$ is negligible.
On the other hand, $w_{\text{SWA}}$ is guaranteed to converge to $\hat w$ 
as $k \rightarrow \infty$.

Moreover, \citet{polyak1992} showed that averaging SGD proposals achieves the best possible
convergence rate among all stochastic gradient algorithms.
The proof relies on the convexity of the underlying problem and
in general there are no convergence guarantees if the loss function is
non-convex \citep[see e.g.][]{ghadimi2013}.
While DNN loss functions are known to be non-convex \citep[e.g.][]{choromanska2015},
over the trajectory of SGD these loss surfaces are approximately convex
\citep[e.g.][]{goodfellow2015}.
However, even when the loss is locally non-convex, SWA can 
improve \emph{generalization}. For example, in Figure \ref{fig:optima_seg} we see that
SWA converges to a central point of the training loss.

In other words, there are a set of points that all achieve low training loss. By running SGD 
with a high constant or cyclical schedule, we traverse over the surface of this set. Then by
averaging the corresponding iterates, we get to move inside the set. This observation explains
both convergence rates and generalization. In deep learning we mostly observe benefits in 
generalization from averaging. Averaging can move to a more central point, which
means one has to move further from this point to increase the loss by a given amount, 
in virtually any direction. By contrast, conventional SGD with a decaying schedule will converge to a point
on the periphery of this set. With different initializations conventional SGD will find different points on the boundary,
of solutions with low training loss, but it will not move inside.

\section{EXPERIMENTS}
\label{sec:experiments}

We compare SWA against conventional SGD training 
on CIFAR-$10$, CIFAR-$100$ and
ImageNet ILSVRC-2012 \citep{russakovsky2015imagenet}.
We also compare to Fast Geometric Ensembling (FGE)
\citep{garipov2018}, but we note that FGE is an ensemble 
whereas SWA corresponds to a single model.  Conventional
SGD training uses a standard decaying learning rate schedule
(details in the Appendix) until convergence. 
We found an exponentially decaying average of 
SGD to perform comparably to conventional SGD at 
convergence. We release the code for reproducing the 
results in this paper at \url{https://github.com/timgaripov/swa}.

\subsection{CIFAR DATASETS}
\label{sec:cifar}

For the experiments on CIFAR datasets we use VGG-16 \citep{simonyan2014very},
a 164-layer Preactivation-ResNet \citep{he2016deep} and Wide
ResNet-28-10 \citep{zagoruyko2016wide} models. Additionally,  we
experiment with the recent Shake-Shake-2x64d \citep{gastaldi2017shake} on CIFAR-$10$
and PyramidNet-$272$ (bottleneck, $\alpha=200$) \citep{han2016deep} on CIFAR-$100$.
All models are trained using $L_2$-regularization, and VGG-$16$ also uses dropout.

For each model we define {\it budget} as the number of epochs required to train
the model until convergence with conventional SGD training, such that we do not
see improvement with SGD beyond this budget.
We use the same budgets for VGG, Preactivation ResNet and Wide ResNet models as 
\citet{garipov2018}. For Shake-Shake and PyramidNets we use the budgets indicated by
the papers that proposed these models \citep{gastaldi2017shake, han2016deep}.
We report the results of SWA training within
$1$, $1.25$ and $1.5$ budgets of epochs.

For VGG, Wide ResNet and Preactivation-ResNet models we first run standard SGD
training for $\approx 75\%$ of the training budget, and then use
the weights at the last epoch as an initialization for SWA with a fixed learning
rate schedule. We ran SWA for $0.25$, $0.5$ and $0.75$ budget to complete
the training within $1$, $1.25$ and $1.5$ budgets respectively.

For Shake-Shake and PyramidNet architectures we do not report the results in one
budget. For these models we use a full budget to get an initialization for the
procedure, and then train with a cyclical learning rate schedule for $0.25$ 
and $0.5$ budgets. We used long cycles of small learning rates for Shake-Shake,
because this architecture already involves many stochastic components.

We present the details of the learning rate schedules for each of these models in 
the Appendix.

For each model we also report the results of conventional SGD training, 
which we denote by \textbf{SGD}. For VGG, Preactivation ResNet and Wide ResNet
we also provide the results of the \textbf{FGE} method with one budget reported in 
\citet{garipov2018}. Note that for FGE we report the accuracy of an ensemble 
of $6$ to $12$ networks, while for SWA we report the accuracy of a single model.

We summarize the experimental results in Table \ref{table:main_experiments}.
For all models we report the mean and standard deviation of test accuracy over $3$ runs. In all
conducted experiments SWA substantially outperforms SGD in one budget, and
improves further, as we allow more training epochs. Across different architectures
we see consistent improvement by $\approx 0.5\%$ on CIFAR-$10$ 
(excluding Shake-Shake, for which SGD performance is already extremely high) and
by $0.75$-$1.5\%$ on CIFAR-$100$. Amazingly, SWA is able to achieve comparable or
better performance than FGE ensembles with just one model. On CIFAR-$100$ SWA
usually needs more than one budget to get results comparable with FGE ensembles, 
but on CIFAR-$10$ even with $1$ budget SWA outperforms FGE.

\subsection{IMAGENET}
\label{sec:imagenet}

On ImageNet we experimented with ResNet-$50$, ResNet-$152$ \citep{he2016deep} and
DenseNet-$161$ \citep{huang2017densely}. For these architectures we used pretrained
models from \href{http://pytorch.org/docs/master/torchvision/models.html}{\texttt{PyTorch.torchvision}}.
For each of the models we ran SWA for $10$ epochs with a cyclical learning rate
schedule with the same parameters for all 
models (the details can be found in the Appendix), and  
report the mean and standard deviation of test error averaged over $3$ runs.
The results are shown in Table~\ref{table:imagenet_experiments}.

\begin{table}[!h]
	\caption{Top-1 accuracies ($\%$) on ImageNet for SWA and SGD with different 
		architectures.}
\label{table:imagenet_experiments}
\centering
\begin{tabular}{cccc}
	\toprule
  & & \multicolumn{2}{c}{SWA}	\\
	\cline{3-4}
	DNN & SGD & $5$ epochs& $10$ epochs \\
	\hline
	ResNet-50 & 
	$76.15$ & $76.83 \pm 0.01$ & $76.97 \pm 0.05$ \\
	ResNet-152 & 
	$78.31$ & $78.82 \pm 0.01$ & $78.94 \pm 0.07$ \\
	DenseNet-161 & 
	$77.65$ & $78.26 \pm 0.09$ & $78.44 \pm 0.06$ \\
	\bottomrule
\end{tabular}
\end{table}

For all $3$ architectures SWA provides consistent improvement by $0.6$-$0.9 \%$
over the pretrained models.

\subsection{EFFECT OF THE LEARNING RATE SCHEDULE}

\begin{figure}[!h]
	\centering
	\includegraphics[width=0.43\textwidth]{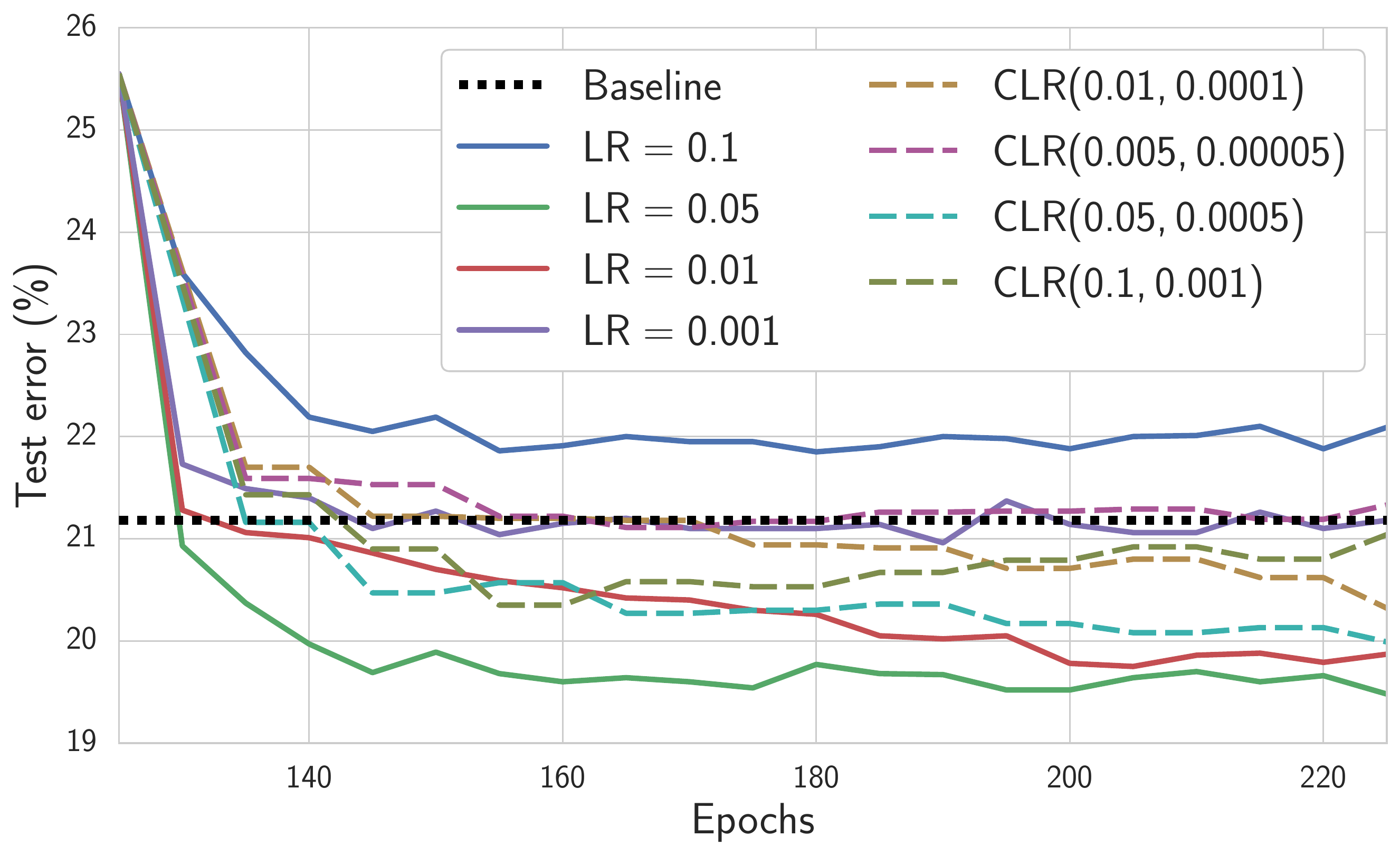}
	\caption{
    Test error as a function of training epoch for SWA with different
    learning rate schedules with a Preactivation ResNet-$164$ on CIFAR-$100$.} 
  \label{fig:lr_dependence}
\end{figure}

In this section we explore how the learning rate schedule affects the 
performance of SWA. We run experiments on Preactivation ResNet-$164$ 
on CIFAR-$100$. For all schedules we use the same initialization from
a model trained for $125$ epochs using the conventional SGD training.
As a baseline we use a fully-trained model trained with conventional
SGD for $150$ epochs.

We consider a range of constant and cyclical learning rate schedules. 
For cyclical learning rates we fix the cycle length to $5$, and consider 
the pairs of base learning rate parameters $(\alpha_1, \alpha_2) \in 
\{(10^{-1}, 10^{-3}), (5 \cdot 10^{-2}, 5 \cdot 10^{-4}), (10^{-2}, 10^{-4}),
(5 \cdot 10^{-3}, 5 \cdot 10^{-5})
\}$. Among the constant learning rates we consider
$\alpha_1 \in \{10^{-1}, 5 \cdot 10^{-2}, 10^{-2}, 10^{-3}\}$. 

We plot the test error of the SWA procedure for different learning rate schedules 
as a function of the number of training epochs in Figure \ref{fig:lr_dependence}.

We find that in general the more aggressive constant learning rate schedule leads
to faster convergence of SWA. In our experiments we found that setting the 
learning rate to some 
intermediate value between the largest and the smallest learning rate
used in the annealing scheme in conventional training usually gave us the best
results.
The approach is however universal and can work well
with different learning rate schedules tailored for particular tasks.

\subsection{DNN TRAINING WITH A FIXED LEARNING RATE}
\label{sec:const_lr}

\begin{figure}[!h]
	\centering
	\includegraphics[width=0.43\textwidth]{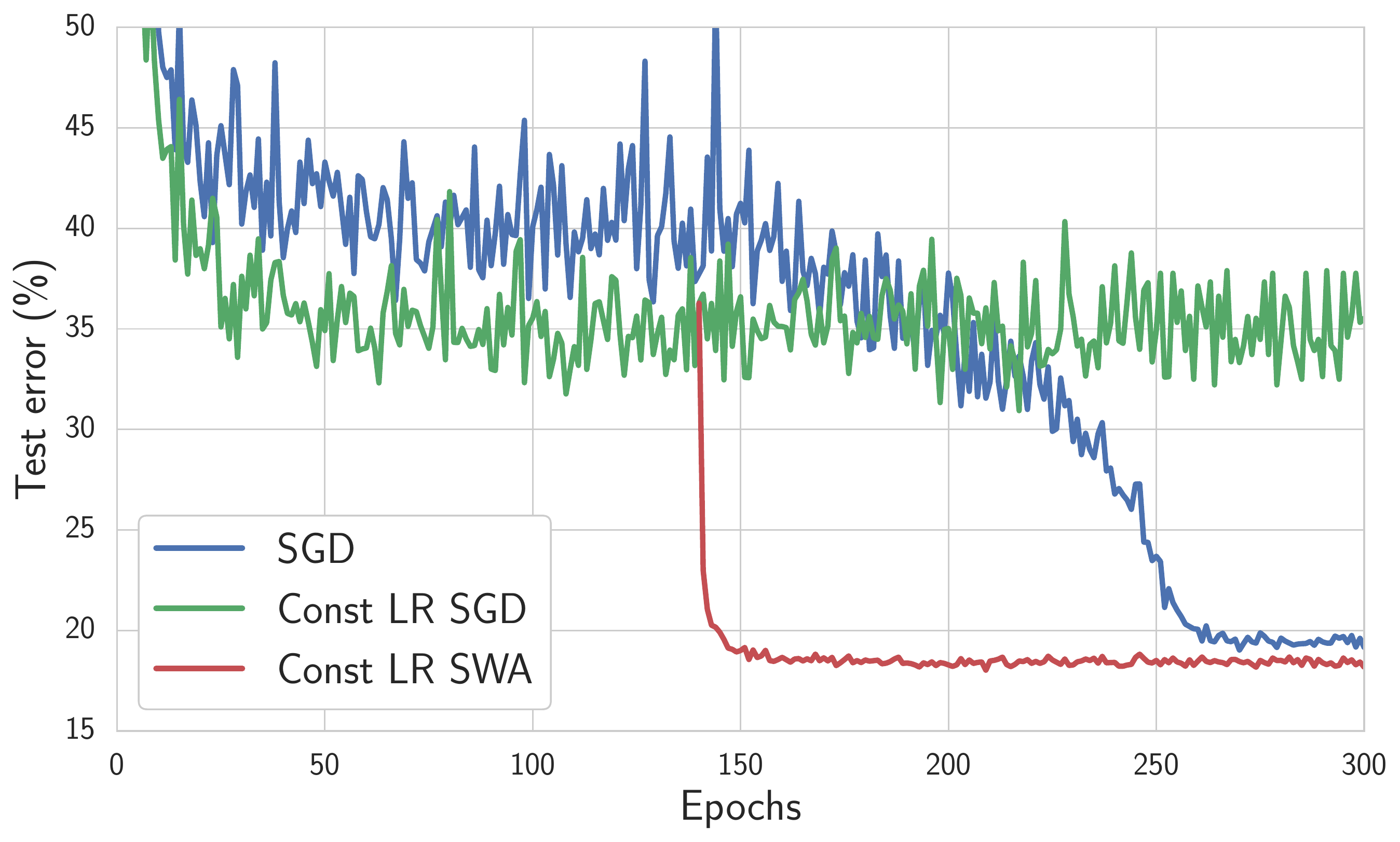}
	\caption{Test error as a function of training epoch for constant (green) and 
  decaying (blue) learning rate schedules for 
  a Wide ResNet-$28$-$10$ on CIFAR-100. In red we average
  the points along the 
  trajectory of SGD with constant learning rate starting at epoch $140$.}
  \label{fig:wrn_convergence}
\end{figure}

In this section we show that it is possible to train DNNs from scratch with a fixed
learning rate using SWA. We run SGD with a fixed learning rate of 
$0.05$ on a Wide ResNet-$28$-$10$ \citep{zagoruyko2016wide} for $300$ epochs from 
a random initialization on CIFAR-100.
We then averaged the weights at the end of
each epoch from epoch $140$ and until the end of training. The final test
accuracy of this SWA model was $81.7$. 

Figure \ref{fig:wrn_convergence} illustrates
the test error as a function of the number of training epochs
for SWA and conventional training. The accuracy of the individual
models with weights averaged by SWA stays at the level of $\approx 65\%$ which is $16 \%$
less than the accuracy of the SWA model.  These results correspond 
to our intuition presented in section \ref{sec:geometry}
that SGD with a constant learning rate oscillates around the 
optimum, but SWA converges.

While being able to train a DNN with a fixed learning rate is a surprising property
of SWA, for practical purposes we recommend initializing SWA from a model
pretrained with conventional training (possibly for a reduced number of epochs),
as it leads to faster and more stable convergence than running SWA from scratch.

\section{DISCUSSION}

We have presented Stochastic Weight Averaging (SWA) for training neural networks. 
SWA is extremely easy to implement, architecture-agnostic, and improves generalization
performance at virtually no additional cost over conventional training.  

There are so many exciting directions for future research. 
SWA does not require each weight in its average to correspond to a good solution, due to the 
geometry of weights traversed by the algorithm.  It therefore may be  
possible to develop SWA for much faster
convergence than standard SGD.  One may also be able to 
combine SWA with large batch sizes while preserving generalization performance, 
since SWA discovers much broader optima than conventional 
SGD training.  Furthermore, a cyclic learning rate enables SWA to
explore regions of high posterior density over neural network weights.
Such learning rate schedules could be developed in conjunction with 
stochastic MCMC approaches, to encourage exploration while still providing 
high quality samples.  One could also develop SWA to average whole regions
of good solutions, using the high-accuracy curves discovered in \citet{garipov2018}.

A better understanding of the loss surfaces for multilayer networks will help continue to 
unlock the potential of these rich models. 
We hope that SWA will inspire further progress in this area.

\paragraph{Acknowledgements.} This work was supported by NSF
IIS-1563887, Samsung Research, Samsung Electronics and
Russian Science Foundation grant 17-11-01027. We also thank Vadim Bereznyuk for 
helpful comments.

\bibliography{bibliography}

\begin{thebibliography}{24}
\providecommand{\natexlab}[1]{#1}
\providecommand{\url}[1]{\texttt{#1}}
\expandafter\ifx\csname urlstyle\endcsname\relax
  \providecommand{\doi}[1]{doi: #1}\else
  \providecommand{\doi}{doi: \begingroup \urlstyle{rm}\Url}\fi

\bibitem[Chaudhari et~al.(2017)Chaudhari, Choromanska, Soatto, LeCun, Baldassi,
  Borgs, Chayes, Sagun, and Zecchina]{chaudhari2016}
P.~Chaudhari, Anna Choromanska, S.~Soatto, Yann LeCun, C.~Baldassi, C.~Borgs,
  J.~Chayes, Levent Sagun, and R.~Zecchina.
\newblock Entropy-sgd: Biasing gradient descent into wide valleys.
\newblock In \emph{International Conference on Learning Representations
  (ICLR)}, 2017.

\bibitem[Choromanska et~al.(2015)Choromanska, Henaff, Mathieu, Arous, and
  LeCun]{choromanska2015}
Anna Choromanska, Mikael Henaff, Michael Mathieu, G{\'e}rard~Ben Arous, and
  Yann LeCun.
\newblock The loss surfaces of multilayer networks.
\newblock In \emph{Artificial Intelligence and Statistics}, pages 192--204,
  2015.

\bibitem[Dinh et~al.(2017)Dinh, Pascanu, Bengio, and Bengio]{dinh2017}
Laurent Dinh, Razvan Pascanu, Samy Bengio, and Yoshua Bengio.
\newblock Sharp minima can generalize for deep nets.
\newblock In \emph{International Conference on Machine Learning}, pages
  1019--1028, 2017.

\bibitem[Draxler et~al.(2018)Draxler, Veschgini, Salmhofer, and
  Hamprecht]{draxler2018}
Felix Draxler, Kambis Veschgini, Manfred Salmhofer, and Fred Hamprecht.
\newblock Essentially no barriers in neural network energy landscape.
\newblock In \emph{Proceedings of the 35th International Conference on Machine
  Learning}, pages 1308--1317, 2018.

\bibitem[Garipov et~al.(2018)Garipov, Izmailov, Podoprikhin, Vetrov, and
  Wilson]{garipov2018}
Timur Garipov, Pavel Izmailov, Dmitrii Podoprikhin, Dmitry~P Vetrov, and
  Andrew~Gordon Wilson.
\newblock Loss surfaces, mode connectivity, and fast ensembling of dnns.
\newblock \emph{arXiv preprint arXiv:1802.10026}, 2018.

\bibitem[Gastaldi(2017)]{gastaldi2017shake}
Xavier Gastaldi.
\newblock Shake-shake regularization.
\newblock \emph{arXiv preprint arXiv:1705.07485}, 2017.

\bibitem[Ghadimi and Lan(2013)]{ghadimi2013}
Saeed Ghadimi and Guanghui Lan.
\newblock Stochastic first-and zeroth-order methods for nonconvex stochastic
  programming.
\newblock \emph{SIAM Journal on Optimization}, 23\penalty0 (4):\penalty0
  2341--2368, 2013.

\bibitem[Goodfellow et~al.(2015)Goodfellow, Vinyals, and Saxe]{goodfellow2015}
Ian~J Goodfellow, Oriol Vinyals, and Andrew~M Saxe.
\newblock Qualitatively characterizing neural network optimization problems.
\newblock \emph{International Conference on Learning Representations}, 2015.

\bibitem[Han et~al.(2016)Han, Kim, and Kim]{han2016deep}
Dongyoon Han, Jiwhan Kim, and Junmo Kim.
\newblock Deep pyramidal residual networks.
\newblock \emph{arXiv preprint arXiv:1610.02915}, 2016.

\bibitem[He et~al.(2016)He, Zhang, Ren, and Sun]{he2016deep}
Kaiming He, Xiangyu Zhang, Shaoqing Ren, and Jian Sun.
\newblock Deep residual learning for image recognition.
\newblock In \emph{Proceedings of the IEEE conference on computer vision and
  pattern recognition}, pages 770--778, 2016.

\bibitem[Hochreiter and Schmidhuber(1997)]{hochreiter1997flat}
Sepp Hochreiter and J{\"u}rgen Schmidhuber.
\newblock Flat minima.
\newblock \emph{Neural Computation}, 9\penalty0 (1):\penalty0 1--42, 1997.

\bibitem[Huang et~al.(2017)Huang, Liu, Weinberger, and van~der
  Maaten]{huang2017densely}
Gao Huang, Zhuang Liu, Kilian~Q Weinberger, and Laurens van~der Maaten.
\newblock Densely connected convolutional networks.
\newblock In \emph{Proceedings of the IEEE conference on computer vision and
  pattern recognition}, volume~1, page~3, 2017.

\bibitem[Ioffe and Szegedy(2015)]{ioffe2015}
Sergey Ioffe and Christian Szegedy.
\newblock Batch normalization: Accelerating deep network training by reducing
  internal covariate shift.
\newblock In \emph{International Conference on Machine Learning}, pages
  448--456, 2015.

\bibitem[Keskar et~al.(2017)Keskar, Mudigere, Nocedal, Smelyanskiy, and
  Tang]{keskar2017large}
Nitish~Shirish Keskar, Dheevatsa Mudigere, Jorge Nocedal, Mikhail Smelyanskiy,
  and Ping Tak~Peter Tang.
\newblock On large-batch training for deep learning: Generalization gap and
  sharp minima.
\newblock \emph{International Conference on Learning Representations}, 2017.

\bibitem[Loshchilov and Hutter(2017)]{loshchilov2016}
Ilya Loshchilov and Frank Hutter.
\newblock Sgdr: stochastic gradient descent with restarts.
\newblock \emph{International Conference on Learning Representations}, 2017.

\bibitem[Mandt et~al.(2017)Mandt, Hoffman, and Blei]{mandt2017stochastic}
Stephan Mandt, Matthew~D Hoffman, and David~M Blei.
\newblock Stochastic gradient descent as approximate bayesian inference.
\newblock \emph{The Journal of Machine Learning Research}, 18\penalty0
  (1):\penalty0 4873--4907, 2017.

\bibitem[Neklyudov et~al.(2018)Neklyudov, Molchanov, Ashukha, and
  Vetrov]{neklyudov2018}
Kirill Neklyudov, Dmitry Molchanov, Arsenii Ashukha, and Dmitry Vetrov.
\newblock Variance networks: When expectation does not meet your expectations.
\newblock \emph{arXiv preprint arXiv:1803.03764}, 2018.

\bibitem[Polyak and Juditsky(1992)]{polyak1992}
Boris~T Polyak and Anatoli~B Juditsky.
\newblock Acceleration of stochastic approximation by averaging.
\newblock \emph{SIAM Journal on Control and Optimization}, 30\penalty0
  (4):\penalty0 838--855, 1992.

\bibitem[Ruppert(1988)]{ruppert1988}
David Ruppert.
\newblock Efficient estimations from a slowly convergent robbins-monro process.
\newblock Technical report, Cornell University Operations Research and
  Industrial Engineering, 1988.

\bibitem[Russakovsky et~al.(2012)Russakovsky, Deng, Su, Krause, Satheesh, Ma,
  Huang, Karpathy, Khosla, Bernstein, et~al.]{russakovsky2015imagenet}
Olga Russakovsky, Jia Deng, Hao Su, Jonathan Krause, Sanjeev Satheesh, Sean Ma,
  Zhiheng Huang, Andrej Karpathy, Aditya Khosla, Michael Bernstein, et~al.
\newblock Imagenet large scale visual recognition challenge.
\newblock \emph{International Journal of Computer Vision}, 115\penalty0
  (3):\penalty0 211--252, 2012.

\bibitem[Simonyan and Zisserman(2014)]{simonyan2014very}
Karen Simonyan and Andrew Zisserman.
\newblock Very deep convolutional networks for large-scale image recognition.
\newblock \emph{arXiv preprint arXiv:1409.1556}, 2014.

\bibitem[Smith and Topin(2017)]{smith2017exploring}
Leslie~N Smith and Nicholay Topin.
\newblock Exploring loss function topology with cyclical learning rates.
\newblock \emph{arXiv preprint arXiv:1702.04283}, 2017.

\bibitem[Srivastava et~al.(2014)Srivastava, Hinton, Krizhevsky, Sutskever, and
  Salakhutdinov]{srivastava2014dropout}
Nitish Srivastava, Geoffrey Hinton, Alex Krizhevsky, Ilya Sutskever, and Ruslan
  Salakhutdinov.
\newblock Dropout: A simple way to prevent neural networks from overfitting.
\newblock \emph{The Journal of Machine Learning Research}, 15\penalty0
  (1):\penalty0 1929--1958, 2014.

\bibitem[Zagoruyko and Komodakis(2016)]{zagoruyko2016wide}
Sergey Zagoruyko and Nikos Komodakis.
\newblock Wide residual networks.
\newblock \emph{arXiv preprint arXiv:1605.07146}, 2016.

\end{thebibliography}
\bibliographystyle{plainnat}
\appendix

\vspace{-2mm}
\section{Appendix}

\subsection{EXPERIMENTAL DETAILS}
For the experiments on CIFAR datasets (section \ref{sec:cifar}) we used the 
following implementations (embedded links):
\begin{itemize}
	\item \href{https://github.com/hysts/pytorch_image_classification}{Shake-Shake-$2$x$64$d}
	\item \href{https://github.com/dyhan0920/PyramidNet-PyTorch}{PyramidNet-$272$}
	\item \href{https://github.com/pytorch/vision/blob/master/torchvision/models/vgg.py}{VGG-$16$}
	\item \href{https://github.com/bearpaw/pytorch-classification/blob/master/models/cifar/preresnet.py}{Preactivation-ResNet-$164$}
	\item \href{https://github.com/meliketoy/wide-resnet.pytorch/blob/master/networks/wide_resnet.py}{Wide ResNet-$28$-$10$}
\end{itemize}

Models for ImageNet are from \href{https://github.com/pytorch/vision/tree/master/torchvision}{here}.
Pretrained networks can be found \href{https://github.com/pytorch/pytorch/blob/master/torch/utils/model_zoo.py}{here}.

\paragraph{SWA learning rates.}
For PyramidNet SWA uses a cyclic learning rate with $\alpha_1 =0.05$ and 
$\alpha_2=0.001$ and cycle length $3$. 
For VGG and Wide ResNet we used constant learning $\alpha_1 = 0.01$. For ResNet 
we used constant learning rates $\alpha_1 = 0.01$ on CIFAR-$10$ and $0.05$ on
CIFAR-$100$. 

\begin{figure}[!t]
	\centering
	\includegraphics[width=0.4\textwidth]{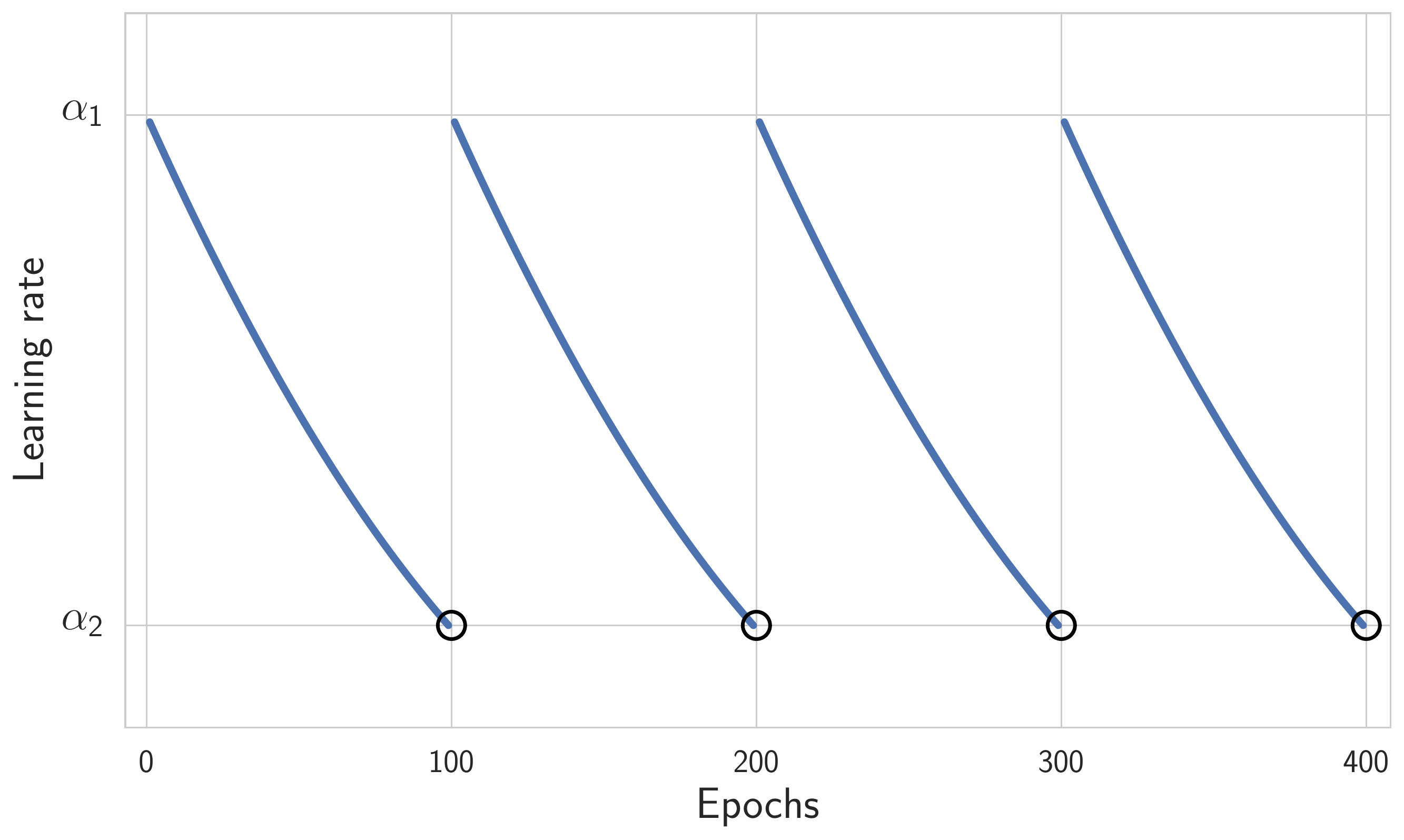}
	\caption{
        Cyclical learning rate used for Shake-Shake as a function of iteration. 
    }
	\label{fig:lr_shk}    	
\end{figure}

For Shake-Shake Net we used a custom cyclic learning rate based on the cosine
annealing used when training Shake-Shake with SGD. Each of the cycles replicate
the learning rates corresponding to epochs $1600-1700$ of the standard training
and the cycle length $c = 100$ epochs. The learning rate schedule is depicted in Figure
\ref{fig:lr_shk} and follows the formula
\begin{align}
  \notag
  \alpha(i) = 0.1 \cdot \left(1 + \cos\left(\pi\cdot \frac{1600 + { \text{epoch}(i)}\bmod{100})}{1800}\right)\right),
\end{align}
where $\text{epoch(i)}$ is the number of data passes completed before iteration
$i$.

For all experiments with ImageNet we used cyclic learning rate schedule with the same 
hyperparameters  $\alpha_1 =0.001$, $\alpha_2=10^{-5}$ and $c=1$.

\paragraph{SGD learning rates.}
For conventional SGD training we used SGD with momentum $0.9$ and with
an annealed learning rate schedule. For VGG, Wide ResNet and Preactivation 
ResNet we fixed the learning rate to $\alpha_1$ for the first half of epochs
($0B$--$0.5B$), then linearly decreased the learning rate to $0.01 \alpha_1$ 
for the next $40\%$ of epochs ($0.5B$--$0.9B$), and then kept it constant
for the last $10\%$ of epochs ($0.9B$ -- $1 B$). For VGG we set $\alpha_1 = 0.05$,
and for Preactivation ResNet and Wide ResNet we set $\alpha_1 = 0.1$. For 
Shake-Shake Net and PyramidNets we used the cosine and piecewise-constant 
learning rate schedules described in \citet{gastaldi2017shake} and \citet{han2016deep}
respectively.

\subsection{TRAINING RESNET WITH A CONSTANT LEARNING RATE}

\begin{figure}[!h]
	\centering
	\includegraphics[width=0.4\textwidth]{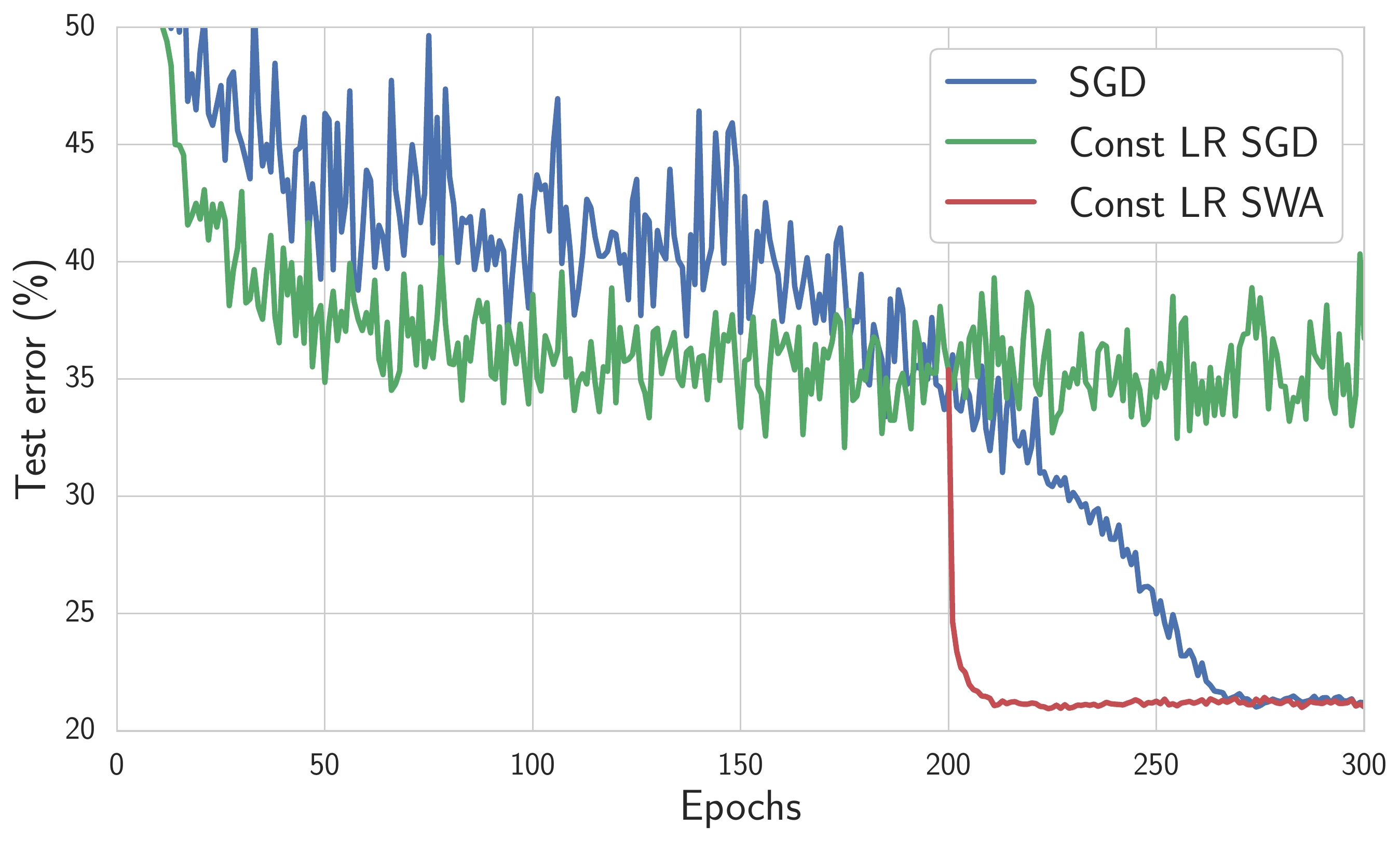}
	\caption{Test error as a function of training epoch for constant (green) and 
  decaying (blue) learning rate schedules for 
  a Preactivation ResNet-$164$ on CIFAR-100. In red we average
  the points along the 
  trajectory of SGD with constant learning rate starting at epoch $200$.}
  \label{fig:resnet_convergence}
\end{figure}

In this section we present the experiment on training Preactivation ResNet-$164$
using a constant learning rate. The experimental setup is the same as in 
section \ref{sec:const_lr}. We set the learning rate to $\alpha_1 = 0.1$ and
start averaging after epoch $200$. The 
results are presented in Figure \ref{fig:resnet_convergence}.

\end{document}